\documentclass[runningheads]{llncs}

\usepackage{eccv}

\usepackage{eccvabbrv}

\usepackage{graphicx}
\usepackage{booktabs}

\usepackage{siunitx}
\usepackage{makecell}
\usepackage{xcolor}
\usepackage{colortbl}

\usepackage[accsupp]{axessibility}  %

\usepackage{hyperref}
\hypersetup{pagebackref,breaklinks,colorlinks,citecolor=eccvblue,
pdftitle={HUE Dataset: High-Resolution Event and Frame Sequences for Low-Light Vision},
pdfsubject={Computer Vision, Event Vision},
pdfauthor={Burak Ercan, Onur Eker, Aykut Erdem, Erkut Erdem},
pdfkeywords={Event cameras, Dynamic Vision Sensor, Event-based Vision, Hybrid Camera System, Low-Light Image Enhancement}
}

\usepackage{orcidlink}

\usepackage{pifont}%

\newcommand\datname{HUE\xspace}
\newcommand\datnamelong{Hacettepe University Event\xspace}

\begin{document}

\title{\datname Dataset: High-Resolution Event and Frame Sequences for Low-Light Vision} 

\titlerunning{HUE Dataset}

\author{Burak Ercan$^*$\inst{1}\orcidlink{0000-0002-9231-7982} \and
Onur Eker$^*$\inst{1,2}\orcidlink{0000-0003-4040-6438} \and
Aykut Erdem\inst{3,4}\orcidlink{0000-0002-6280-8422} \and
Erkut Erdem\inst{1}\orcidlink{0000-0002-6744-8614}}

\authorrunning{B.~Ercan et al.}

\institute{Hacettepe University, Computer Engineering Department \and
HAVELSAN Inc. \and
Ko\c{c} University, Computer Engineering Department \and
Ko\c{c} University, KUIS AI Center
}

\maketitle
\def\thefootnote{*}\footnotetext{These authors contributed equally to this work}

\begin{abstract}

Low-light environments pose significant challenges for image enhancement methods. To address these challenges, in this work, we introduce the HUE dataset, a comprehensive collection of high-resolution event and frame sequences captured in diverse and challenging low-light conditions. Our dataset includes 106 sequences, encompassing indoor, cityscape, twilight, night, driving, and controlled scenarios, each carefully recorded to address various illumination levels and dynamic ranges. Utilizing a hybrid RGB and event camera setup. we collect a dataset that combines high-resolution event data with complementary frame data. We employ both qualitative and quantitative evaluations using no-reference metrics to assess state-of-the-art low-light enhancement and event-based image reconstruction methods. Additionally, we evaluate these methods on a downstream object detection task. Our findings reveal that while event-based methods perform well in specific metrics, they may produce false positives in practical applications. This dataset and our comprehensive analysis provide valuable insights for future research in low-light vision and hybrid camera systems.  
\end{abstract}

\section{Introduction}
\label{sec:intro}

The growing interest in event-based vision has led to the development of numerous datasets that capture dynamic scenes with high temporal resolution and a wide dynamic range. However, existing datasets often come with certain limitations, such as low resolution, limited diversity of scenes, or a lack of challenging low-light conditions. Given these constraints, our motivation for presenting the \datname (\datnamelong) dataset is threefold. First, our dataset captures events at $1280 \times 720$, surpassing the resolution of most available datasets. High-resolution event data is crucial for accurately capturing fine details in complex scenes, enabling more precise analysis and reconstruction tasks. Second, it features a substantial collection of sequences shot in diverse settings. Specifically, the  \datname dataset includes 106 sequences recorded in both indoor and outdoor environments, using cameras that are either handheld or mounted on a vehicle. These sequences are captured at various times of the day, such as sunset, twilight, and nighttime, and include a range of camera motions from slow to fast. The scenes also feature both static and dynamic objects, including people, animals, vehicles, buildings, everyday objects, cityscapes, and landscapes. This variety ensures that the dataset can be used to evaluate a wide array of methods under different conditions. Third, the \datname dataset specifically targets challenging low-light scenarios, with sequences captured in conditions where the illuminance on the event sensor is just a few lux. This focus on low-light environments addresses a significant gap in current datasets, which often do not adequately represent such conditions, and is of utmost importance for critical applications like automotive~\cite{shariff2024event}, flight control~\cite{vitale2021event}, and robotics~\cite{yang2023neuromorphic}.

Our dataset is collected using a setup consisting of two cameras: one event camera and one frame camera. The event sequences, therefore, include complementary frames as well. We employ a non-coaxial setup where the two cameras do not share the same optical axis and have different optical characteristics. Consequently, the event data and frames collected are not pixel-wise spatially aligned. 
We share an additional calibration sequence with a flickering chessboard pattern so that camera calibration procedures can be applied to estimate intrinsic and extrinsic camera parameters.
Non-coaxial datasets like ours offer the advantage of simpler collection methods, as they do not require additional equipment such as optical beam splitters or specialized cameras like DAVIS~\cite{brandli2014240}. This generic setup is anticipated to become increasingly popular due to its ease of integration, such as the multi-camera systems commonly found in smartphones and robotic systems~\cite{wang2021stereo}. Consequently, methods targeting unaligned event and frame data are being developed as well~\cite{cho2023non}, in contrast to earlier works that make the limiting assumption of pixel-wise alignment.

Using the low-light scenes from our dataset, we conduct a comprehensive evaluation of state-of-the-art low-light image enhancement~\cite{cai2023retinexformer,xu2022snr,zhang2021learning} and event-based
image reconstruction~\cite{weng2021event,ercan2024hypere2vid} and enhancement \cite{liang2023coherent} methods, where we qualitatively and quantitatively compare and contrast their advantages and shortcomings.

In summary, the main contributions of our work are as follows:
\begin{itemize}
    \item We introduce the \datname dataset, which offers high-resolution event data and complementary frame data captured in diverse and challenging low-light scenarios.
    \item We provide a comprehensive evaluation of state-of-the-art low-light enhancement and event-based image reconstruction methods using our dataset.
    \item We highlight the importance of color information in practical applications, particularly in downstream tasks like object detection.
\end{itemize}

The rest of the paper is organized as follows: Section 2 discusses related work, including comparable low-light image enhancement datasets and event-based vision datasets. Section 3 details our data collection setup and the characteristics of the HUE dataset. Section 4 presents our experimental evaluation of various methods on the dataset. Finally, Section 5 concludes the paper and discusses future directions. Our dataset, additional calibration sequence, and evaluation scripts are available at \url{https://ercanburak.github.io/HUE.html}.

\section{Related Work}

\subsection{Datasets for Event-Guided Low-light Enhancement}

Collecting paired low-light/normal-light RGB and event data is challenging, resulting in very few real-world datasets. Jiang \etal~\cite{jiang2023event} introduced the LIE dataset, the first real-world dataset designed specifically for this purpose. It includes event streams and underexposed frames captured in various indoor and outdoor settings, mostly below 2 lux, using the DAVIS346 event camera for accurate photometric data. The dataset was collected by adjusting the camera’s light intake in static scenes, triggering events with light changes indoors and different exposure times outdoors.
Liang \etal~\cite{liang2023coherent} evaluated their methods on real-world data using a custom hybrid camera system. This system combines an industrial camera (FLIR Chameleon 3 Color, $1920\times1080$ at 20 FPS) and an event camera (DAVIS346, $346\times260$) using a beam splitter mounted in front of the two cameras with 50\% optical splitting.
The EvLight framework, proposed by Liang \etal~\cite{liang2024towards}, introduces a new dataset with over 30K pairs of spatially and temporally aligned images and events under various lighting conditions. This dataset was captured using a robotic arm to ensure precise alignment. The robotic system follows a predefined path, and the DAVIS346 event camera operates with fixed parameters like exposure time. Initially, paired image and event sequences are captured under normal lighting conditions. An ND8 filter is then applied to the camera lens to capture low-light sequences while maintaining consistent camera settings.
There are also event datasets that include night sequences without specifically targeting the task of event-guided low-light enhancement. However, these suffer from limitations such as very few night sequences and low resolution~\cite{scheerlinck2018continuous,zhu2018multivehicle}, lack of complementary frames~\cite{perot2020learning}, or limited diversity of the scenes~\cite{gehrig2021dsec}.

\renewcommand{\arraystretch}{1.15}
\begin{table}[!b]
\centering
\caption{Real-world datasets collected for event-guided low-light enhancement.}
\resizebox{\textwidth}{!}{
\begin{tabular}{@{}c@{$\;\;\;$}c@{$\;\;\;$}c@{$\;\;\;$}c@{$\;\;\;$}c@{$\;\;\;$}c@{$\;\;\;$}c@{$\;\;\;$}c@{$\;\;\;$}c@{}}
\toprule
\textbf{Dataset} & \textbf{Lux} & \textbf{Event Res.} & \textbf{Frame Res.} & \textbf{Seqs.} & \textbf{GT} & \textbf{Release} & \textbf{Scene} & \textbf{Camera} \\ \midrule
\cite{jiang2023event} & 0.1-5 & 346\(\times\)260 & 346\(\times\)260 & 206 & \ding{51} & \ding{55} & Static & Static \\ 
\cite{liang2024towards} & - & 346\(\times\)260 & 346\(\times\)260 & 91 & \ding{51} & \ding{51} & Static & Dynamic \\ 
\cite{liang2023coherent} & - & 346\(\times\)260 & 1920\(\times\)1280 & - & \ding{55} & \ding{55} & - & - \\ 
Ours & 0-24 & 1280\(\times\)720 & 1456\(\times\)1088 & 106 & \ding{55} & \ding{51} & Dynamic & Dynamic \\ \bottomrule
\end{tabular}}
\label{tab:ll_event_datasets}
\end{table}

Table~\ref{tab:ll_event_datasets} highlights the advantages of our dataset for event-guided low-light enhancement compared to existing datasets targeting this task. One key benefit is our dataset's much higher resolution, with events captured at $1280\times720$ and frames at $1456\times1088$. This provides more detailed and high-quality data for both events and frames. In contrast, other datasets, such as those by Jiang \etal~\cite{jiang2023event} and Liang \etal~\cite{liang2024towards}, use much lower resolutions of 346×260 for both events and frames. Even though Liang \etal~\cite{liang2023coherent} achieves a frame resolution of $1920\times1280$, their event resolution is still much lower. Moreover, our dataset includes dynamic scenes captured with moving cameras, offering a wider range of real-world applications. This is a significant improvement over other methods that typically use static scenes or stationary cameras.

\subsection{Low-light Enhancement}
Enhancing images and videos captured under low-light conditions is a significant challenge in computer vision. Traditional low-light enhancement techniques have largely focused on histogram equalization~\cite{pizer1987adaptive,pizer1990contrast,ibrahim2007brightness,arici2009histogram} and Retinex theory-based methods~\cite{jobson1997properties,wang2013naturalness,wang2014variational,guo2016lime,cai2017joint} to improve image quality. Recently, deep learning approaches have provided more robust and adaptive solutions. Additionally, event-guided methods have further expanded the potential for low-light enhancement by utilizing the high dynamic range and temporal resolution of event cameras, overcoming the limitations of conventional frame-based methods.

\noindent \textbf{Low-light image enhancement.} The first deep learning-based LLIE method, LLNet~\cite{lore2017llnet}, employs a variant of stacked-sparse denoising autoencoder to simultaneously brighten and denoise low-light images. Lv \etal~\cite{lv2018mbllen} introduced an end-to-end multi-branch enhancement network (MBLLEN) that includes a feature extraction module, an enhancement module, and a fusion module to extract effective feature representations. Xu \etal~\cite{xu2022snr} proposed using the signal-to-noise ratio (SNR) as prior information for enhancing low-light images, employing convolutional neural networks (CNNs) to encode local information in high SNR regions and transformers to capture relationships with distant pixels in noisy low SNR regions. Wu \etal~\cite{wu2023learning} addressed color consistency issues in enhanced dark images by using a semantic segmentation method to produce and utilize semantic priors, resulting in higher color consistency. Wei \etal\cite{wei2018deep} combined Retinex theory with CNNs to develop RetinexNet, which effectively enhances low-light images. Cai \etal~\cite{cai2023retinexformer} proposed the Retinex-based Illumination-Guided Transformer (IGT), which uses illumination representations to guide self-attention computation and enhance interactions between regions of different exposure levels.

\noindent \textbf{Low-light video enhancement.} Chen \etal~\cite{chen2019seeing} proposed a siamese network to ensure generalization capability for processing videos of dynamic scenes. Wang \etal~\cite{wang2021seeing} formulated an end-to-end framework for enhancing underexposed videos, emphasizing a self-supervised strategy for noise reduction and illumination enhancement based on Retinex theory. To enhance moving objects in dark videos, Haiyang \etal~\cite{jiang2019learning} proposed a 3D U-Net-based network to map RAW low-light videos to normal-light videos. Zhang \etal~\cite{zhang2021learning} introduced a method that enforces temporal stability in low-light video enhancement by inferring motion fields from static images and synthesizing short-range video sequences. Fu \etal~\cite{fu2023dancing} addressed camera motion and spatial alignment by proposing a Retinex-based Light Adjustable Network (LAN) for iterative refinement and adaptive illumination adjustment.

\noindent \textbf{Event-guided low-light enhancement.} Liu \etal~\cite{liu2023low} introduced a low-light video enhancement approach using synthetic event guidance, featuring an Event and Image Fusion Transform (EIFT) module for event-image fusion and an Event-Guided Dual Branch (EGDB) module for low-light enhancement, encompassing event synthesis, event and image fusion, and low-light enhancement. Jiang \etal~\cite{jiang2023event} utilized a residual fusion module to integrate features from event and frame data, addressing the domain gap with a feature pyramid structure and multi-level encoder for high-order semantic information extraction. Liang \etal~\cite{liang2023coherent} proposed a novel approach with a multimodal coherence modeling module to establish coherence between frames and events, and a temporal coherence propagation module for coherence-aware aggregation. The EvLight framework\cite{liang2024towards} featured an event-guided low-light enhancement approach with SNR-guided regional feature selection to fuse high-SNR image regions with events from low-SNR regions, and a holistic-regional fusion branch to integrate structural and textural information from both events and images.

\subsection{Event-based Image Reconstruction}

Reconstructing intensity images from event data is a key task in event-based vision, bridging image and event domains. This allows visualization of events, application of frame-based methods to event data~\cite{rebecq2019events,muglikar2021calibrate}, and making use of relations between these two domains in contexts like unsupervised domain adaptation~\cite{sun2022ess,xie2024cross,zheng2024eventdance}. While earlier works rely on some limiting assumptions such as known or restricted camera movement, static scenes, or brightness constancy \cite{kim2014simultaneous,kim2016real,bardow2016simultaneous}, recent deep learning based methods trained on large synthetic datasets show impressive results without such assumptions \cite{rebecq2019high,stoffregen2020reducing,weng2021event,ercan2024hypere2vid}. Zhang \etal~\cite{zhang2020learning} specifically target event data of low-light scenes and propose an unsupervised domain adaptation network to generate intensity images as if captured in daylight. However, their codes or models are not publicly shared. We use two recent open-source event-based image reconstruction methods for our experimental analysis, ET-Net~\cite{weng2021event} and HyperE2VID~\cite{ercan2024hypere2vid}, which can successfully reconstruct fine details in dark scenes, where traditional frame cameras suffer from underexposure or noise. The details of these two methods are presented in \Cref{sec:exp}. 

\section{Dataset}

\subsection{Data Collection Setup}
Different types of optical setups can be used to collect complementary event and frame data. One option is to use a camera that integrates both event-based and frame-based data collection mechanisms within the same pixel array, such as the Dynamic and Active Pixel Vision Sensor (DAVIS)~\cite{brandli2014240}. This setup provides synchronized and pixel-wise aligned events and frames. However, it has two significant drawbacks: the shutter activity associated with each frame introduces noise into the event data, which is absent in sensors dedicated solely to event detection, and the most advanced model of this hybrid camera, the DAVIS346, only supports a maximum resolution of 346 × 260 pixels.
The second option involves using a beam splitter, an optical device featuring a 50/50 mirror that reflects half of the incoming light and transmits the other half. This arrangement allows frame and event sensors to share an aligned field of view. Despite its advantages, this setup requires additional equipment, which incurs extra costs and occupies more space.
The third option, which we have chosen, is a stereo hybrid event-frame camera setup, where separate event and frame cameras are mounted side by side. This setup offers simplicity, as it does not require a beam splitter or a specialized low-resolution camera like the DAVIS. The details of our data collection setup are presented in the following.

\subsubsection{Hardware and Optics}

Our setup utilizes a PROPHESEE Gen4M event camera~\cite{finateu20205} and an Allied Vision Alvium compact CMOS camera. The event camera features a Sony IMX636 event-based sensor, which is in a $1/2.5''$ format with pixel dimensions of $4.86~\mu m \times 4.86~\mu m$. It offers a resolution of $1280 \times 720$ and a dynamic range exceeding $120$ dB. The event camera is equipped with a Soyo SFA 0820-5M lens, having a fixed focal length of $8$ mm, a maximum aperture of $f/2.0$, a minimum focus distance of $0.1$ meters, and a horizontal field of view of approximately $38^{\circ}$.
The RGB camera contains a Sony IMX273 global shutter sensor, which is in a $1/2.9''$ format with pixel dimensions of $3.45~\mu m \times 3.45~\mu m$. It provides a resolution of $1456 \times 1088$, a dynamic range of $75$ dB, and a $12$-bit analog-to-digital converter. The RGB camera is paired with a Tamron M118FM06 lens, having a fixed focal length of $6$ mm, a maximum aperture of $f/1.4$, a minimum focus distance of $0.1$ meters, and a horizontal field of view of approximately $45^{\circ}$.

\begin{figure}[tb]
\centering
\includegraphics[width=0.4\textwidth]{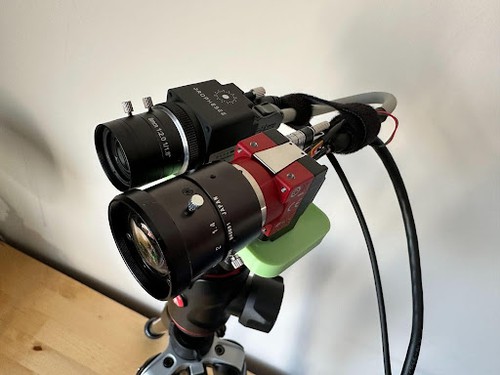}
\caption{A picture showing our data collection setup containing a PROPHESEE Gen4M event camera paired with an Allied Vision Alvium compact traditional CMOS camera.}
\vspace{-2mm}
\label{fig:rwdc_hw}
\end{figure}

The cameras are positioned with a baseline distance of approximately 2 cm between their optical axes, and their relative positions are kept fixed in an optical setup. To achieve this, a custom mechanical part designed specifically for the cameras was produced using a 3D printer. A picture of our setup is presented in Fig.~\ref{fig:rwdc_hw}. 
To capture objects at varying distances in focus, both lenses have been set to a narrow aperture of $f/8$, providing a wide depth of field. The focusing distances of both lenses were then adjusted to the hyperfocal distance, ensuring sharp capture of all objects beyond a certain distance, such as $0.5$ meters. For time synchronization, the RGB and event cameras are directly connected using a dedicated cable, linking one of the general-purpose input/output (GPIO) lines of the RGB camera to one of the external trigger lines of the event camera. The software details of time synchronization are described in the following section.

\subsubsection{Camera Settings and Software}
The RGB camera is configured to record at a frame rate of 25 frames per second, with an exposure time set to minimize motion blur in dynamic scenes. Specifically, the exposure time is set to a maximum of 35 milliseconds. The digital gain value of the RGB camera is adjusted based on the scene's brightness—higher for dark scenes and lower for bright scenes—to manage image brightness and noise levels. To avoid excessive noise, we prefer slightly underexposed images in low-light scenarios. The PROPHESEE Gen4 event sensor has six basic operating parameters: \texttt{bias\_diff}, \texttt{bias\_diff\_on}, \texttt{bias\_diff\_off}, \texttt{bias\_hpf}, \texttt{bias\_fo}, and \texttt{bias\_refr}. These parameters control the sensor's sensitivity to light intensity changes, the cutoff frequencies of event filters, and the refractory period of each pixel. Adjusting these settings affects the sensor's overall sensitivity, temporal precision, background noise, and event delay. To maintain a balance among these characteristics, we have decided to keep the event sensor's parameters at their factory default settings.

We developed software to complement our hardware setup, running on a computer and communicating with the RGB and event cameras via USB interfaces. This software configures the cameras, controls data acquisition, and stores the acquired events and frames. The image frames from the RGB camera and event streams from the event camera are saved to the computer's permanent memory.
For time synchronization, the GPIO line of the RGB camera is programmed to be at a high voltage level (logic 1) during exposure and at a low voltage level (logic 0) at other times. The event camera continuously monitors the voltage level on its external trigger line and records the transitions with high temporal precision. Consequently, the start and end times of each RGB camera frame exposure are timestamped using the event camera's high-resolution clock, enabling precise synchronization of data from both cameras.
The frames captured by the RGB camera are saved as 
8-bit 3-channel color images. The event camera records brightness change events, including pixel position, polarity (increase or decrease in brightness), and a high-precision timestamp. Additionally, the amount of light falling on the event sensor, measured in lux, is recorded with each frame.

\subsection{Collected Dataset} \label{sec:dataset_splits}
Our dataset includes a diverse array of sequences captured in various challenging scenarios. Comprising 106 sequences with an average duration of approximately 17 seconds, the dataset totals around 30 minutes of footage. This includes over 44,000 frames and approximately 27 billion events. To manage its size and variability, we have divided the dataset into six categories based on the scene characteristics. Below, we explain each category in detail, while \Cref{tab:dat_cat} presents an overview and \cref{fig:dataset_samples} displays frames and event visualizations.

\newcommand*{\theadCustom}[1]{\multicolumn{1}{l}{\textbf{#1}}}
\newcolumntype{R}[1]{>{\RaggedLeft\arraybackslash}p{#1}}
\newcolumntype{C}[1]{>{\centering\arraybackslash}p{#1}}

\setlength{\tabcolsep}{0.2cm} %
\renewcommand{\arraystretch}{1}
\begin{table}[t]
\centering
\caption{Types of scenes in our \datname Dataset.}
\label{tab:dat_cat}
\begin{tabular}{ p{1.2cm} p{0.4cm} p{1.2cm} p{0.8cm} p{5.2cm} }
\toprule
	\theadCustom{{Category}}
	& \theadCustom{{Seqs.}}
    & \theadCustom{{Duration}}
	& \theadCustom{{Lux}}
	& \theadCustom{{Description}} \\     

	\midrule
	Indoor
	& 23
	& 6 mins
	& 0-3
	& Indoor environments dimly lit via natural or artificial light sources. \\[-0.5mm]
 
	\arrayrulecolor{black!30}
    \midrule
 
	City
	& 11
	& 4 mins
	& 0-14
	& Capturing cityscapes through the window of a mid-rise building. \\[-0.5mm]
 
	\midrule
    Twilight
	& 23 	
	& 10 mins
	& 0-24
	& Outdoors during twilight, featuring natural and urban elements. \\[-0.5mm]
 
	\midrule
    Night
	& 16 	
	& 3 mins
	& $\sim$~0
	& Taken in urban environments during night, with moving people, cars, etc. \\[-0.5mm]
 
	\midrule
	Driving 		
	& 16	
	& 6	mins
	& $\sim$~0
	& From the windshield of a car driving around the city in twilight and night. \\[-0.5mm]
 
	\midrule
	Controlled 	
	& 17 	
	& 2 mins   	
	& 0-2
	& Same scene captured under varying lighting levels and camera settings. \\
	\arrayrulecolor{black} 
        \bottomrule
\end{tabular}
\vspace{-2mm}
\end{table}
\renewcommand{\arraystretch}{1}
\setlength{\tabcolsep}{0.05cm} %

\begin{figure}[t]
\small
\newcommand{\widthplot}{0.15\linewidth}
\centering
\setlength{\tabcolsep}{0.4ex} %
\begin{tabular}{ccccccc}

    & Indoor & City & Twilight & Night & Driving & Controlled \\  
    \rotatebox[origin=l]{90}{~Frames} &
    \includegraphics[width=\widthplot]{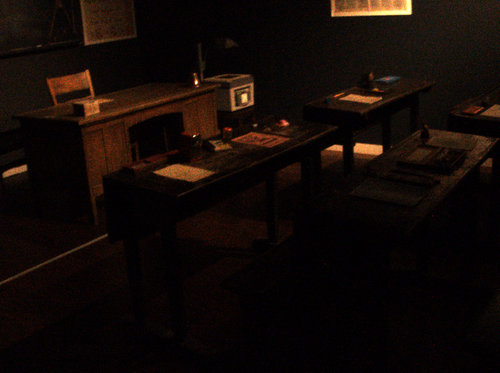} & 
    \includegraphics[width=\widthplot]{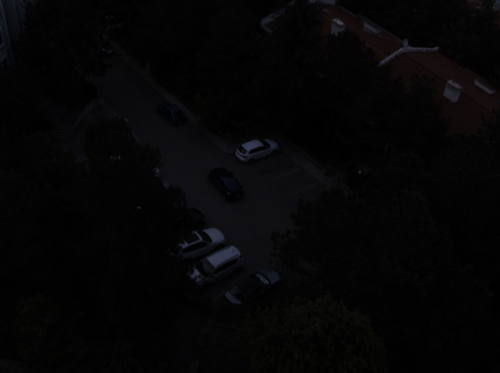} &     
    \includegraphics[width=\widthplot]{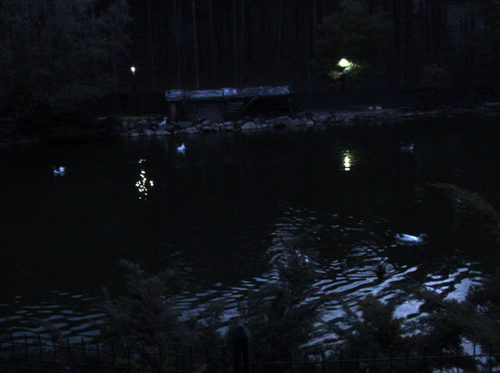} & 
    \includegraphics[width=\widthplot]{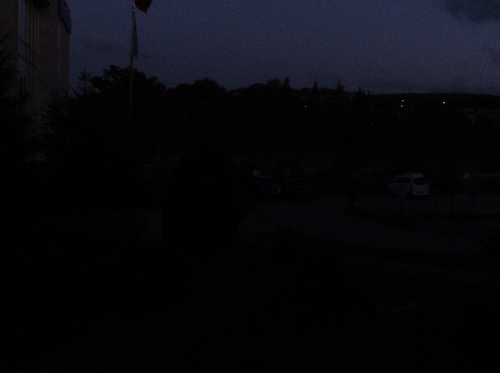} & 
    \includegraphics[width=\widthplot]{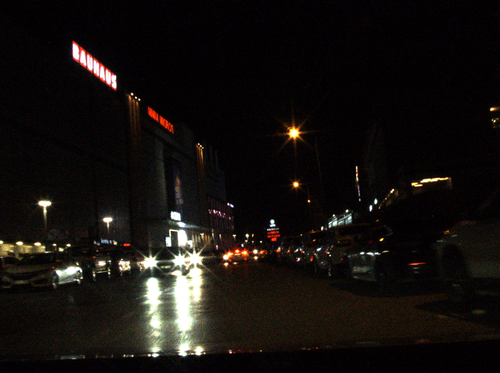} & 
    \includegraphics[width=\widthplot]{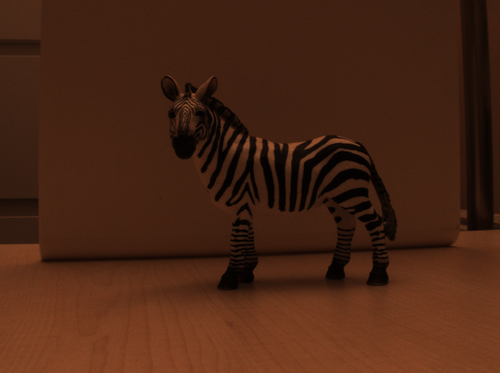} \\[-1ex]
    \rotatebox[origin=l]{90}{Events} &
    \includegraphics[width=\widthplot]{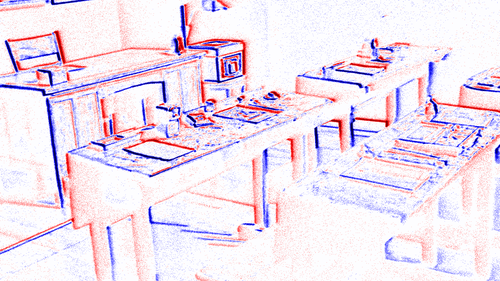} & 
    \includegraphics[width=\widthplot]{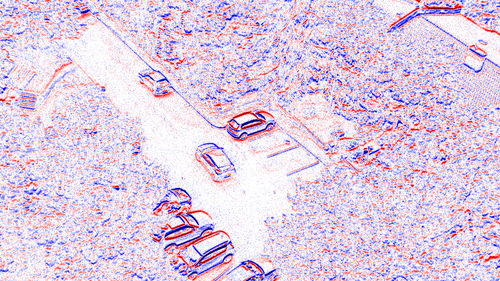} &     
    \includegraphics[width=\widthplot]{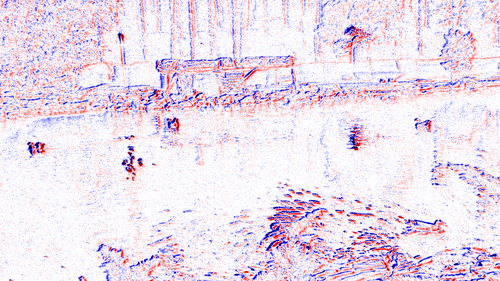} & 
    \includegraphics[width=\widthplot]{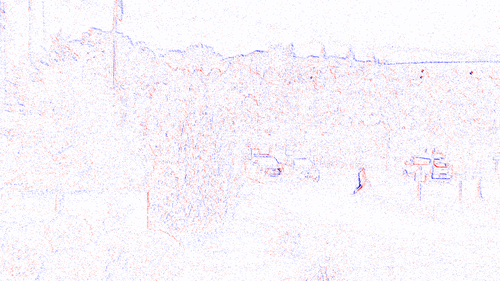} & 
    \includegraphics[width=\widthplot]{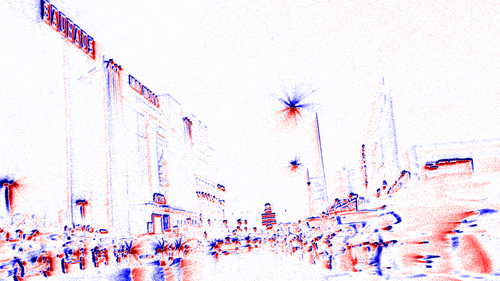} & 
    \includegraphics[width=\widthplot]{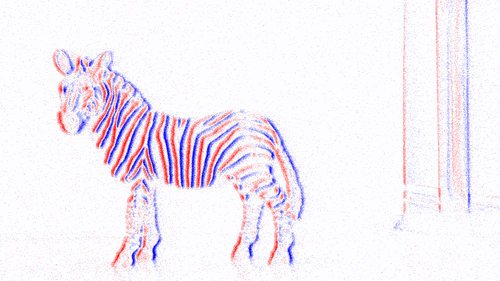} \\
    \rotatebox[origin=l]{90}{~Frames} &
    \includegraphics[width=\widthplot]{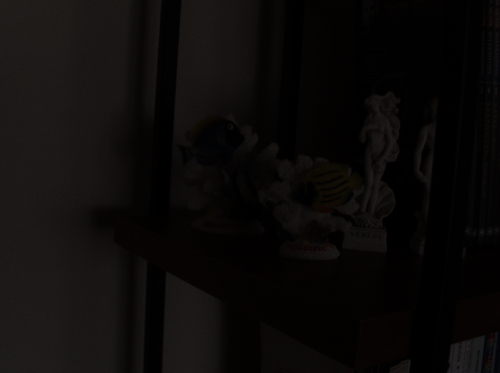} & 
    \includegraphics[width=\widthplot]{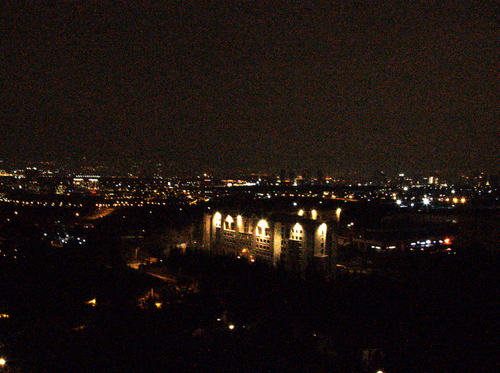} &     
    \includegraphics[width=\widthplot]{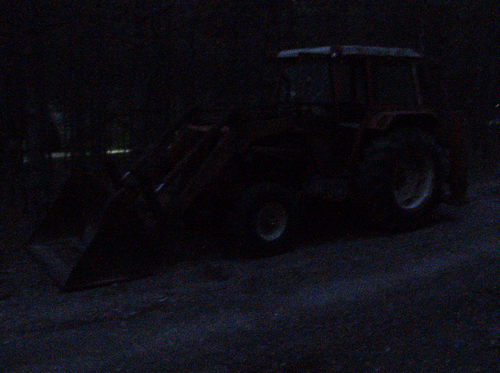} & 
    \includegraphics[width=\widthplot]{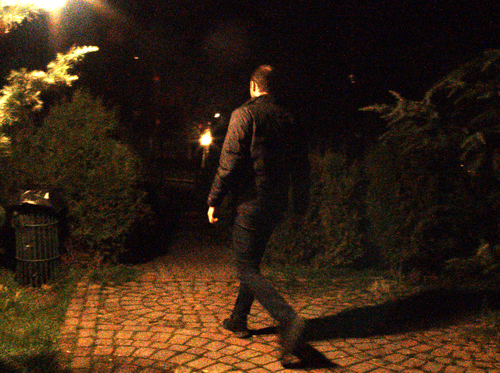} & 
    \includegraphics[width=\widthplot]{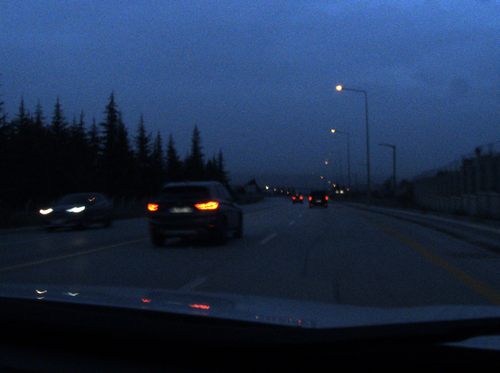} & 
    \includegraphics[width=\widthplot]{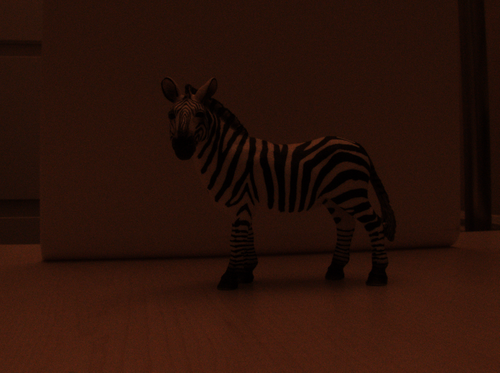} \\ [-1ex]
    \rotatebox[origin=l]{90}{Events} &
    \includegraphics[width=\widthplot]{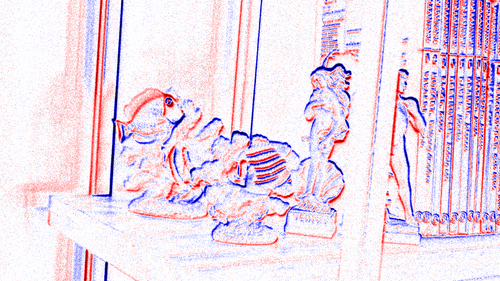} & 
    \includegraphics[width=\widthplot]{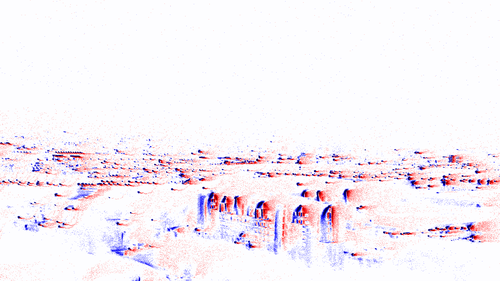} &     
    \includegraphics[width=\widthplot]{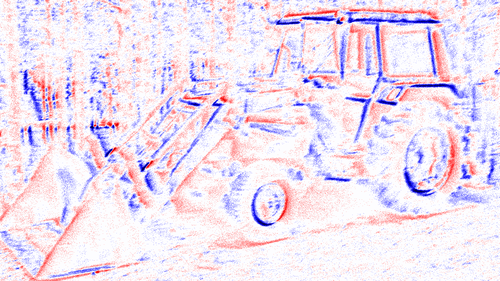} & 
    \includegraphics[width=\widthplot]{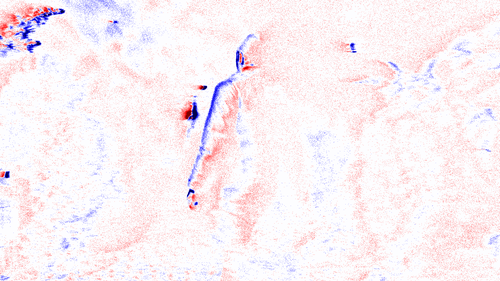} & 
    \includegraphics[width=\widthplot]{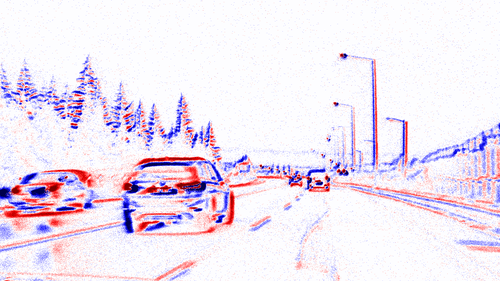} & 
    \includegraphics[width=\widthplot]{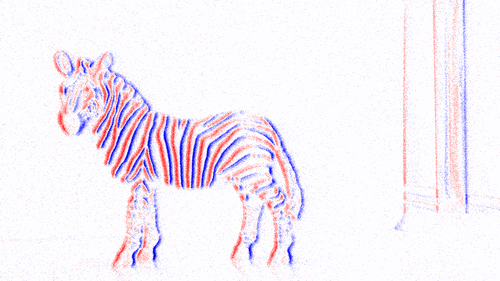} \\ 
    \rotatebox[origin=l]{90}{~Frames} &
    \includegraphics[width=\widthplot]{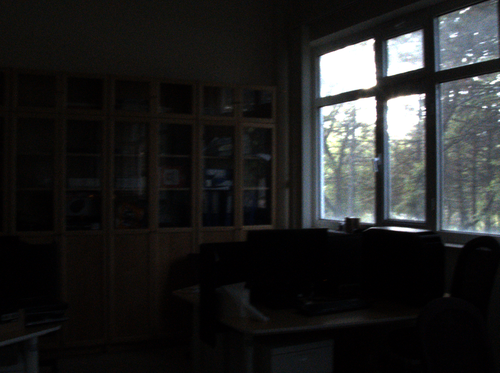} & 
    \includegraphics[width=\widthplot]{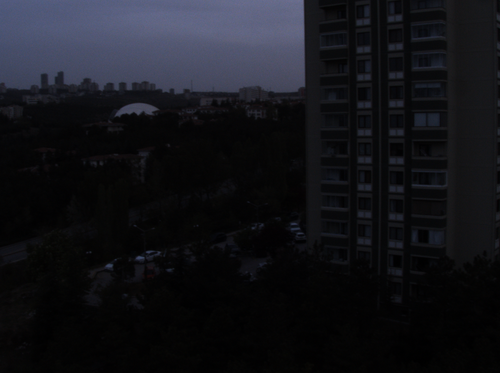} &     
    \includegraphics[width=\widthplot]{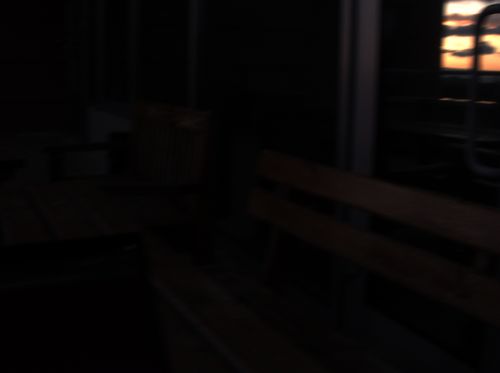} & 
    \includegraphics[width=\widthplot]{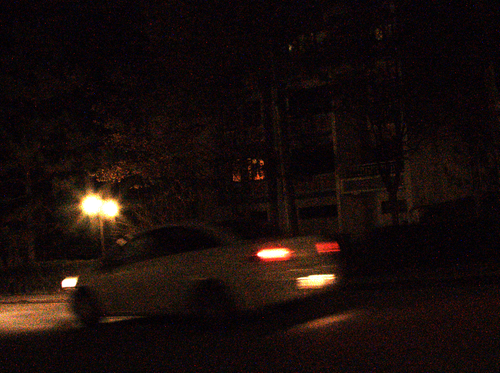} & 
    \includegraphics[width=\widthplot]{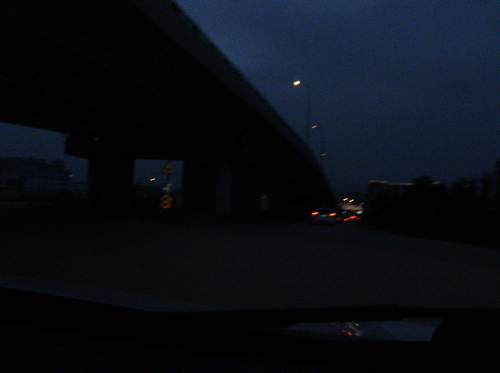} & 
    \includegraphics[width=\widthplot]{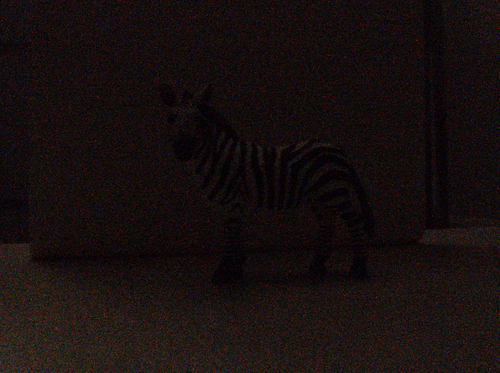} \\ [-1ex]
    \rotatebox[origin=l]{90}{Events} &
    \includegraphics[width=\widthplot]{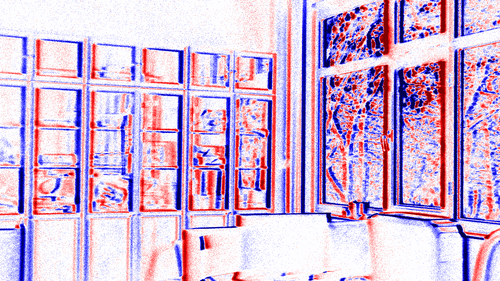} & 
    \includegraphics[width=\widthplot]{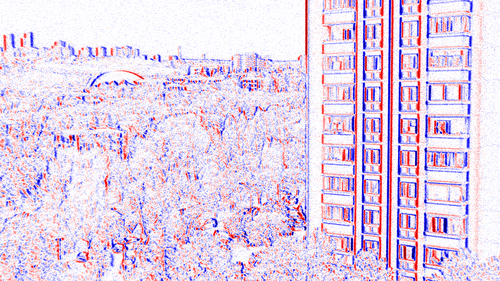} &     
    \includegraphics[width=\widthplot]{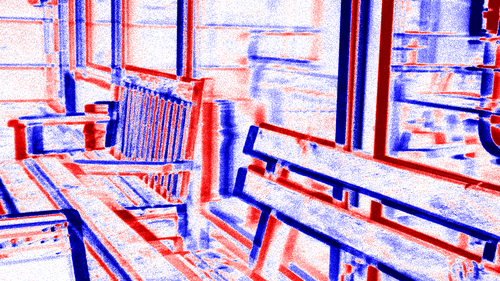} & 
    \includegraphics[width=\widthplot]{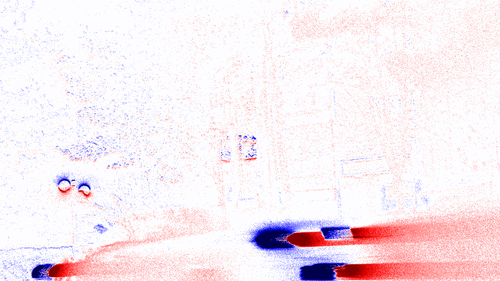} & 
    \includegraphics[width=\widthplot]{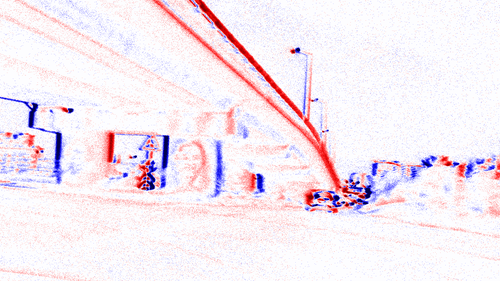} & 
    \includegraphics[width=\widthplot]{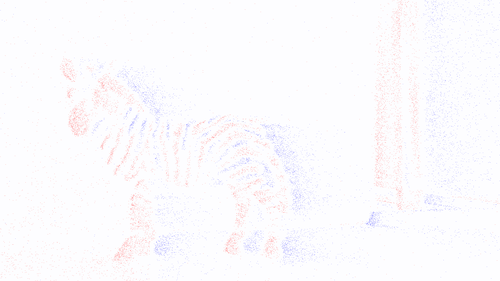} \\
\end{tabular}
\caption{Sample scenes from different categories of our proposed \datname dataset.}
\vspace{-2mm}
\label{fig:dataset_samples}
\end{figure}

\vspace{0.15cm}
\noindent \textbf{\datname-Indoor:} 
This category includes 23 sequences captured in dimly lit indoor environments, with sensor illuminance levels of just a few lux. This setting allows the evaluation of methods under low-light conditions within indoor environments. 14 of these sequences are lit with natural light filtering through windows, while the remaining sequences are illuminated by artificial light sources. Approximately one-third of the sequences contain dynamic scenes. While some sequences are filmed in large indoor halls and corridors featuring distant objects, most focus on closer subjects. \datname-Indoor is particularly valuable for assessing image enhancement performance on objects with fine details, such as small text and textured regions, under low-light conditions.

\vspace{0.15cm}
\noindent \textbf{\datname-City:} 
This category comprises 11 sequences recorded from the window of a mid-rise building, capturing cityscapes. Seven sequences are recorded during twilight hours, while two are captured at night. In these sequences, the camera setup moves slowly, and the scenes are predominantly static, although occasional moving objects such as vehicles or birds are present. Common elements like roads and buildings are mostly distant from the camera, with scenes extending towards the horizon. This setup tests the spatial resolution limits of the event camera and evaluates each method's capability to reconstruct fine details, as the scenes contain many objects and textures represented by just a few pixels on the sensor. The remaining two sequences are taken during the daytime, capturing both indoor and outdoor scenes. These sequences test the performance under high-dynamic range, since the ratio between the illuminance levels of highly lit and low-lit parts of the scene often varies by orders of tens or hundreds.

\vspace{0.15cm}
\noindent \textbf{\datname-Twilight:} 
This portion of our dataset includes 23 sequences taken outdoors during the twilight hours of the evening when the sun is below the horizon at varying degrees. More specifically, we define twilight hours as the period when the solar elevation angle (the angle of the sun's geometric center relative to the horizon) is between $0$ and $-12$ degrees, encompassing both civil twilight and nautical twilight. The primary source of illumination is sunlight scattering in the atmosphere, supplemented occasionally by artificial lights such as vehicle headlamps and streetlights. The average illuminance on the event sensor is just a few lux for most of these scenes. Approximately one third of the sequences feature static scenes, while the remaining majority are dynamic, capturing movement within the environment. The sequences feature a mix of natural and urban elements, ranging from lakes and forests to vehicles and buildings.

\vspace{0.15cm}
\noindent \textbf{\datname-Night:} 
This subset contains 16 sequences, all of which are taken in urban environments during the night, \ie when the solar elevation angle is below $-18$ degrees. This leads to significantly darker conditions, with sensor illumination always being below 1 lux. For these sequences, artificial lights such as streetlights and vehicle headlamps become the primary source of illumination. Half of the sequences feature dynamic elements such as moving people and vehicles. Containing some of the darkest sequences in \datname, this subset is particularly valuable for assessing performance in very low light conditions.

\vspace{0.15cm}
\noindent \textbf{\datname-Driving:} 
This category includes 16 driving sequences, where the camera setup is mounted inside of a vehicle's front windshield, monitoring ahead through this window. Throughout these recordings, the vehicle travels through various street settings in twilight and nighttime, capturing elements such as other cars, pedestrians, motorbikes, parked vehicles, gas stations, tunnels, interchanges, and roundabouts. This subset is important for evaluating performance in dynamic scenes and under challenging lighting conditions, such as dark roads and rapidly moving headlights.

\vspace{0.15cm}
\noindent \textbf{\datname-Controlled:} In this category, we present 17 sequences capturing the same scene, under varying lighting levels and camera settings. More specifically, we record a toy zebra on a desk indoors from a close distance. The recordings are made at night, with the only light source being a dimmable halogen lamp, indirectly illuminating the subject. Throughout these sequences, we adjust the lamp to produce ten different illuminance levels. We record two sequences for each of the first seven illuminance levels, by setting two different digital gain values for the RGB camera. For each of the last three illuminance levels, which are the darkest, we record a single sequence with a gain value of 48, which is the maximum allowed by the RGB camera. This category allows us to make more controlled comparisons under various illuminance levels.

\section{Experiments} \label{sec:exp}

\subsection{Tested Approaches}
We evaluate and compare several state-of-the-art methods belonging to four different categories both quantitatively and qualitatively on our \datname dataset: \textbf{(I)} two RGB image-based methods, RetinexFormer~\cite{cai2023retinexformer} and SNR-Net~\cite{xu2022snr}, \textbf{(II)} an RGB video-based method, namely StableLLVE~\cite{zhang2021learning}, \textbf{(III)} two event-based image reconstruction methods, ET-Net~\cite{weng2021event} and HyperE2VID~\cite{ercan2024hypere2vid}, and \textbf{(IV)} a~hybrid event and RGB video-based method, EvLowLight~\cite{liang2023coherent}.\\

\noindent\textbf{RetinexFormer\cite{cai2023retinexformer}} introduces a one-stage Retinex-based Framework (ORF), revising the traditional Retinex model with perturbation terms for reflectance and illumination to handle corruptions. ORF estimates illumination to brighten images and employs a corruption restorer to reduce noise, artifacts, and color distortions. Additionally, an Illumination-Guided Transformer (IGT) with Illumination-Guided Multi-head Self-Attention (IG-MSA) models long-range dependencies. IGT is integrated into ORF, forming the complete RetinexFormer method.\\

\noindent\textbf{SNR-Net\cite{xu2022snr}} presents an SNR-Aware framework that dynamically adjusts enhancement based on local signal-to-noise ratio (SNR) estimations. It distinguishes between regions with low and high SNR, applying long-range operations to heavily corrupted areas and short-range operations to clearer regions. It includes a noise estimation module and an SNR-aware transformer, selectively using tokens with sufficient SNR to dynamically enhance image quality.\\

\noindent\textbf{StableLLVE\cite{zhang2021learning}} aims to ensure temporal consistency in low-light video enhancement using static images. The core idea is to predict motion fields from a single image to synthesize short-range video sequences, addressing temporal stability challenges. By using optical flow to simulate motion, this approach applies temporal consistency even when training on static images.\\

\noindent\textbf{ET-Net\cite{weng2021event}} combines the strengths of CNNs and Transformers in a hybrid network designed for event-based video reconstruction. It leverages CNNs for capturing fine-scale local analysis and Transformers for global context modeling. ET-Net introduces a Token Pyramid Aggregation (TPA) module to integrate multi-scale token information, enhancing the relation of internal and intersected semantic concepts. This addresses the limitations of CNNs, particularly in modeling long-range dependencies essential for high-quality video reconstruction from event data.\\

\noindent\textbf{HyperE2VID\cite{ercan2024hypere2vid}} introduces a dynamic neural network architecture for event-based video reconstruction. It utilizes hypernetworks to generate per-pixel adaptive filters, informed by a context fusion module that combines data from event voxel grids and previously reconstructed intensity images. HyperE2VID employs a recurrent encoder-decoder backbone with dynamic convolutions applied at the decoder, adapting to spatial variations and enhancing both static and dynamic parts of the scene.\\

\noindent\textbf{EvLowLight\cite{liang2023coherent}} integrates event data with frame data to enhance low-light videos. It leverages the unique ability of event cameras to record brightness changes at extremely high temporal resolution and with a high dynamic range. EvLowLight features a multimodal coherence modeling module that compensates for differences between events and frames, ensuring accurate alignment under challenging low-light conditions. A temporal coherence propagation module further refines the output by sampling features from corresponding points in consecutive events and frames, reducing noise and enhancing the signal-to-noise ratio of the final video output.

\subsection{Evaluation Results}
All comparison results were generated using the official codes, pre-trained models, and default parameters provided for each method. In particular, for RetinexFormer and SNR-Net, we used pre-trained models on the LOL-v1 dataset~\cite{wei2018deep}. StableLLVE was evaluated with the pre-trained model on their custom synthetic dataset. For ET-Net and HyperE2VID, we utilize official pre-trained models trained on a synthetic training set as described in~\cite{stoffregen2020reducing} and employ EVREAL evaluation framework~\cite{ercan2023evreal} to generate reconstructions. EvLowLight was tested using the model trained on their proposed synthetic dataset.

To quantitatively evaluate the quality of the low-light enhancement and event-based image reconstruction methods, we used no-reference metrics since our dataset lacks paired normal-light ground truth images. These metrics provide perceptual quality scores by directly processing the input images without needing ground truth references. We employed four no-reference metrics: BRISQUE~\cite{mittal2012no}, NIQE~\cite{mittal2012making}, MANIQA~\cite{yang2022maniqa}, and MUSIQ~\cite{ke2021musiq}. BRISQUE and NIQE are traditional metrics that use hand-crafted features to measure adherence to natural scene statistics, considering distortions such as blur, noise, and compression. Conversely, MANIQA and MUSIQ are deep-learning-based methods designed to assess perceptual image quality end-to-end, specifically focusing on distortions typically observed in the outputs of neural network-based image restoration algorithms.

Table~\ref{tab:quant_results} presents the quantitative results of the evaluated low-light enhancement and event-based image reconstruction methods on our \datname dataset. It is important to note that these results were obtained without aligning the RGB data with the event data. Hence, the methods were evaluated on slightly different scenes with different resolutions, which may contribute to the variability in their performance.

\begin{table*}[!t]
\centering
\caption{Quantitative results of the methods on no-reference metrics.}
\begin{tabular}{l@{$\quad$}c@{$\quad$}c@{$\quad$}c@{$\quad$}c}
\hline
Methods & NIQE$\;\downarrow$ & BRISQUE$\;\downarrow$ & MANIQA$\;\uparrow$ & MUSIQ$\;\uparrow$ \\
\hline
RetinexFormer\cite{cai2023retinexformer} & \underline{4.56} & \textbf{16.34} & 0.17 & 24.54 \\
SNR-Net\cite{xu2022snr} & 11.07 & 58.63 & 0.18 & 23.71 \\[0.15cm]
StableLLVE\cite{zhang2021learning} & 5.52 & 20.26 & 0.17 & 21.04 \\[0.15cm]
ET-Net\cite{weng2021event} & \textbf{4.21} & 25.62 & \textbf{0.31} & \textbf{46.00} \\
HyperE2VID\cite{ercan2024hypere2vid} & 4.86 & \underline{21.14} & \underline{0.28} & \underline{40.29} \\[0.15cm]
EvLowLight\cite{liang2023coherent} & 6.26 & 47.59 & 0.18 & 37.78 \\
\hline
\end{tabular}
\vspace{-2mm}
\label{tab:quant_results}
\end{table*}

In terms of BRISQUE score,  RetinexFormer achieves the best results, closely followed by the event-based image reconstruction method HyperE2VID. For NIQE, ET-Net leads with the best scores, followed by RetinexFormer, while SNR-Net records the lowest scores for both BRISQUE and NIQE metrics. When considering MANIQA and MUSIQ, the event-based image reconstruction methods ET-Net and HyperE2VID dominate with the highest scores. EvLowLight, a hybrid event and image method, ranks third in MUSIQ, while it shares the third position in MANIQA with the RGB image-only method SNR-Net. StableLLVE shows the lowest score on MUSIQ, and along with RetinexFormer, it has the lowest scores on MANIQA. 

From these results, it is evident that event-based image reconstruction methods generally outperform others in NIQE, MANIQA and MUSIQ metrics, while BRISQUE favors RetinexFormer. Grayscale reconstructions produced by event-based image reconstruction methods are perceived as having less noise and richer texture, contributing to their higher scores in these metrics. Moreover, EvLowLight consistently scores better than other RGB-based methods across metrics except NIQE and BRISQUE. It should be noted that the no-reference metrics used were not specifically designed for low-light enhancement tasks, making their use as direct performance indicators somewhat questionable. These metrics do not adequately account for color and semantic information in their score assessments.

\begin{figure*}[!t]
\scriptsize
\newcommand{\widthframes}{0.12\textwidth}
\newcommand{\widthevents}{0.16\textwidth}
\centering
\setlength{\tabcolsep}{0.4ex} %
\begin{tabular}{cccccccc}
    \rotatebox[origin=l]{90}{Indoor} &

    \includegraphics[width=\widthframes]{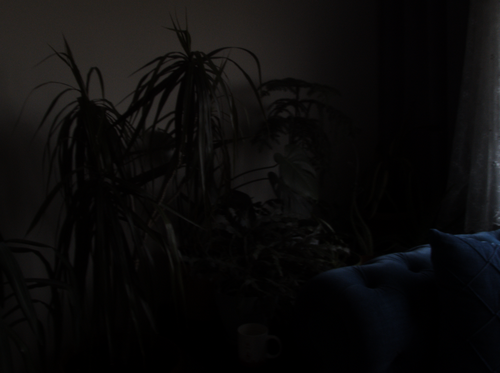} &
	\includegraphics[width=\widthframes]{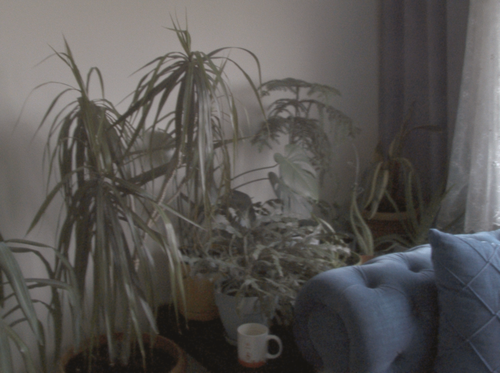} &
	\includegraphics[width=\widthframes]{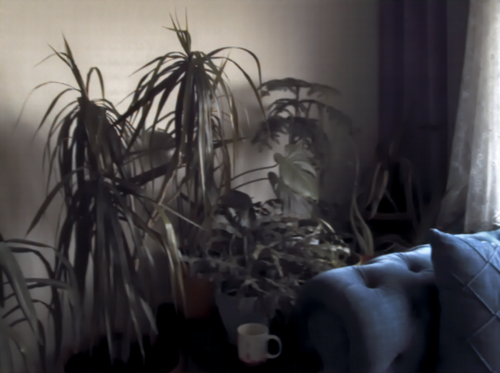} &
    \includegraphics[width=\widthframes]{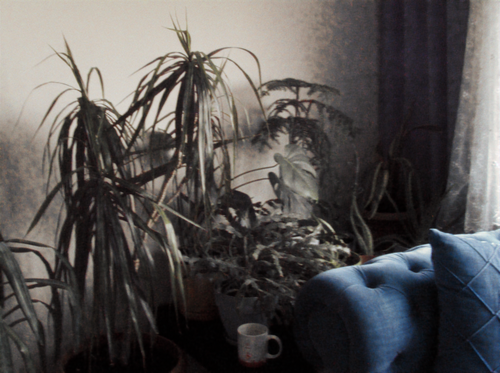} &
    \includegraphics[width=\widthframes]{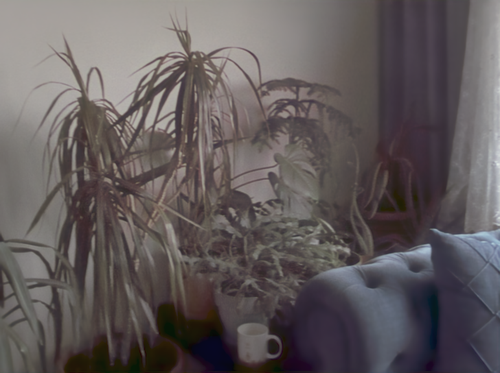} &
	    \includegraphics[width=\widthevents]{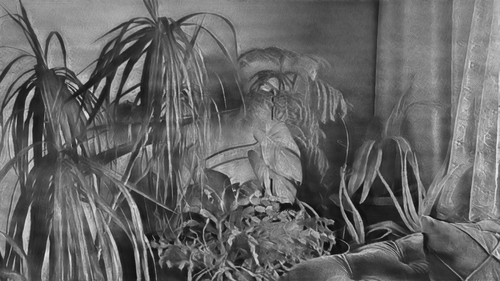} &
	\includegraphics[width=\widthevents]{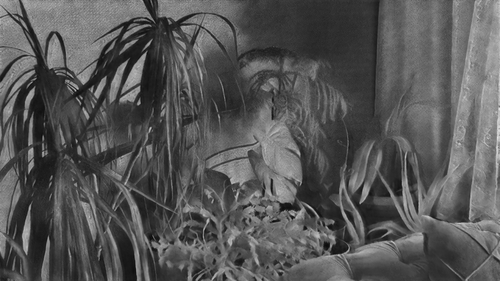} \\	

	\rotatebox[origin=l]{90}{Window} &
    \includegraphics[width=\widthframes]{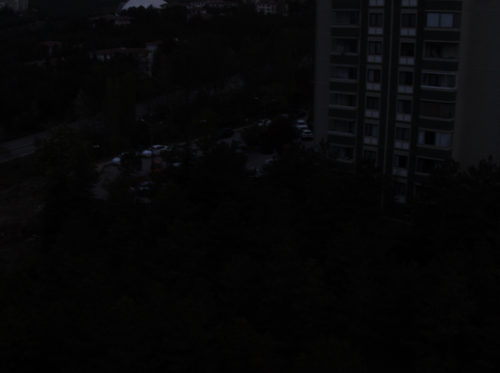} &
	\includegraphics[width=\widthframes]{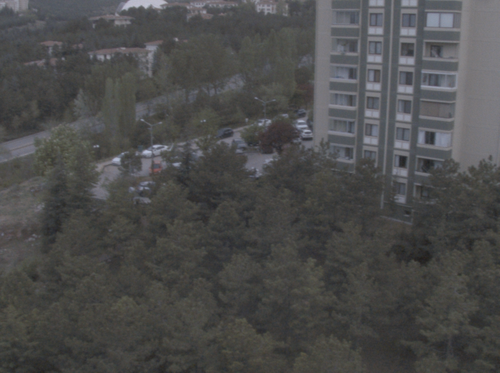} &
	\includegraphics[width=\widthframes]{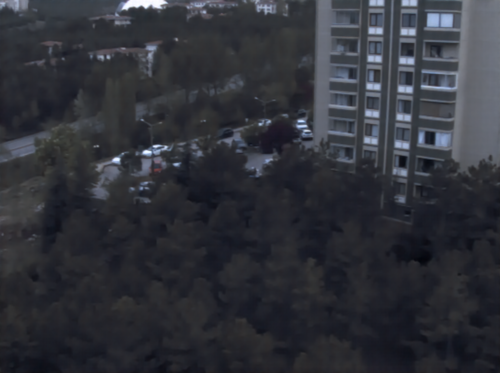} &
	\includegraphics[width=\widthframes]{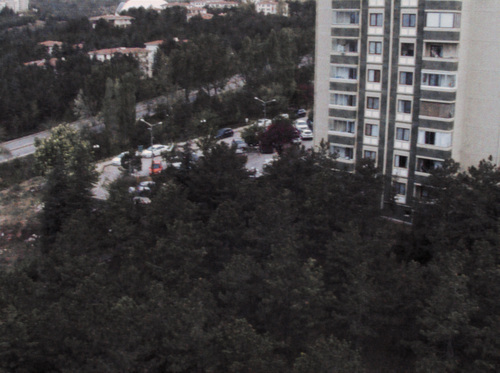} &
	\includegraphics[width=\widthframes]{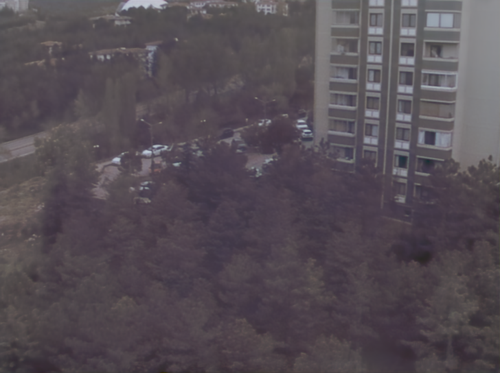} 	&
	\includegraphics[width=\widthevents]{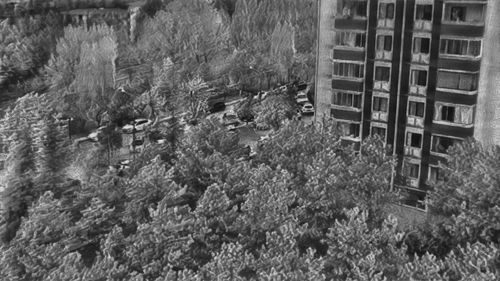} &
	\includegraphics[width=\widthevents]{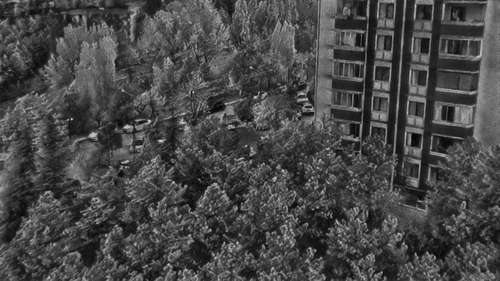} \\

	\rotatebox[origin=l]{90}{Driving} &
    \includegraphics[width=\widthframes]{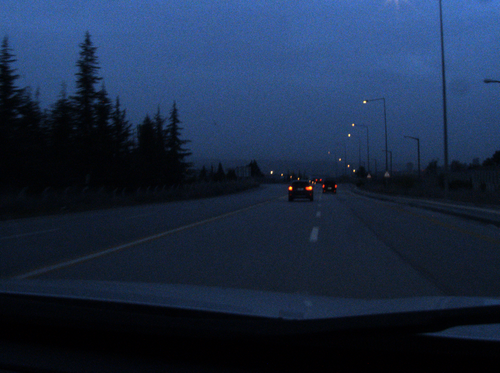} &
	\includegraphics[width=\widthframes]{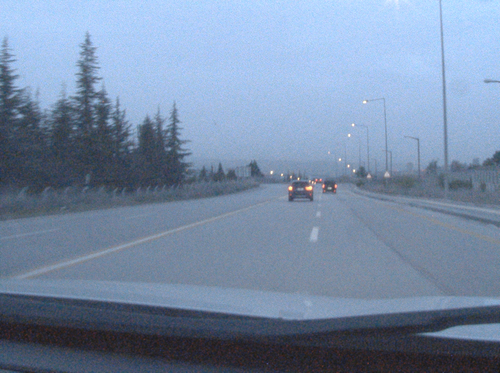} &
	\includegraphics[width=\widthframes]{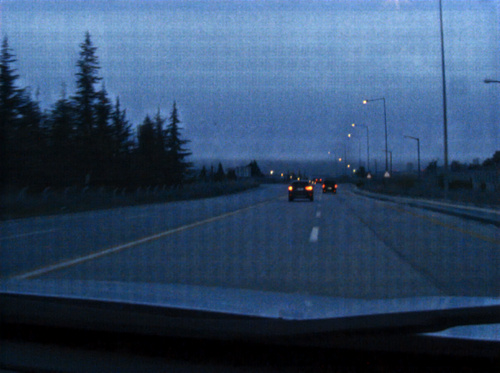} &
	\includegraphics[width=\widthframes]{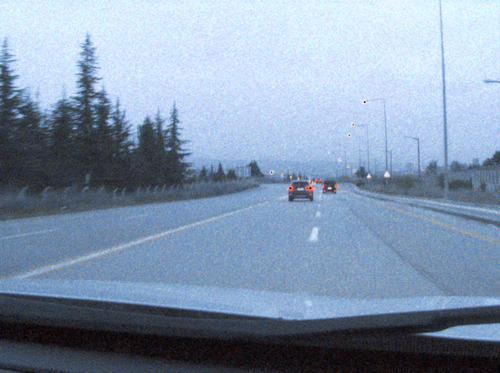} &
	\includegraphics[width=\widthframes]{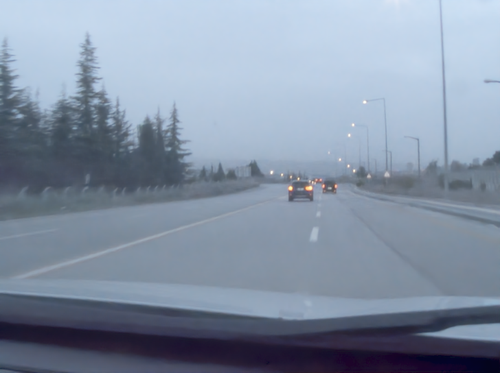} &
    \includegraphics[width=\widthevents]{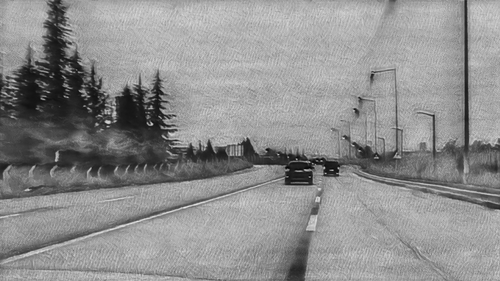} &
	\includegraphics[width=\widthevents]{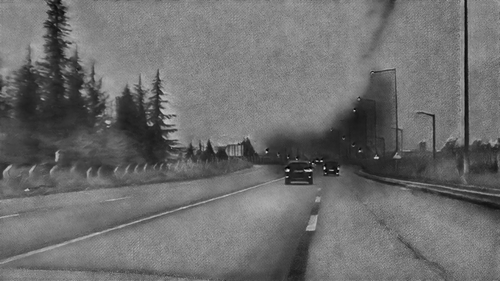} \\

    \rotatebox[origin=l]{90}{Twilight} &
    \includegraphics[width=\widthframes]{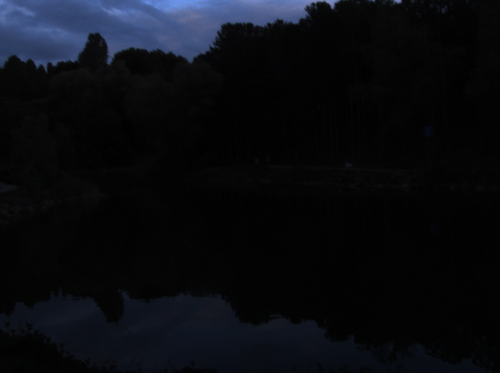} &
	\includegraphics[width=\widthframes]{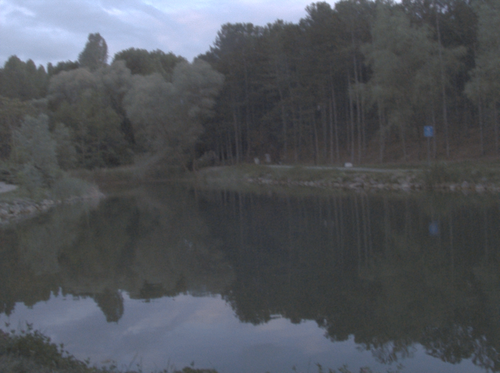} &
	\includegraphics[width=\widthframes]{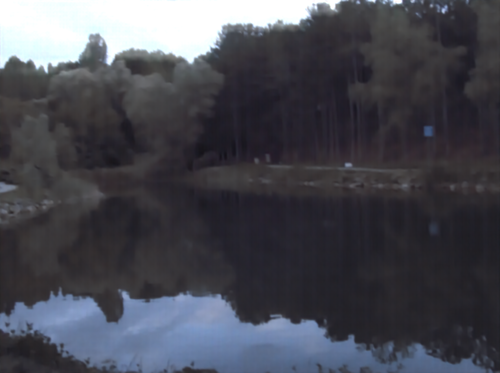} &
	\includegraphics[width=\widthframes]{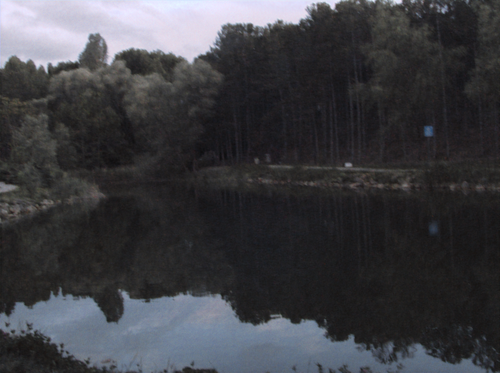} &
	\includegraphics[width=\widthframes]{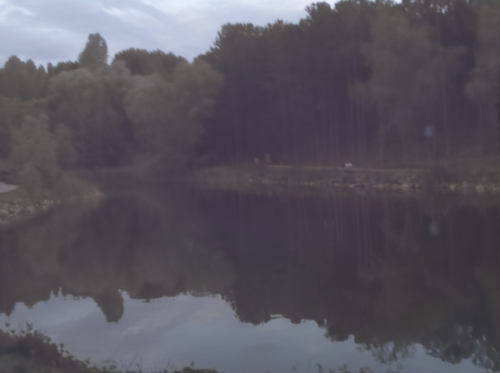} &
    \includegraphics[width=\widthevents]{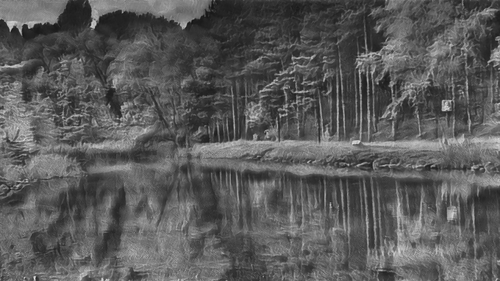} &
	\includegraphics[width=\widthevents]{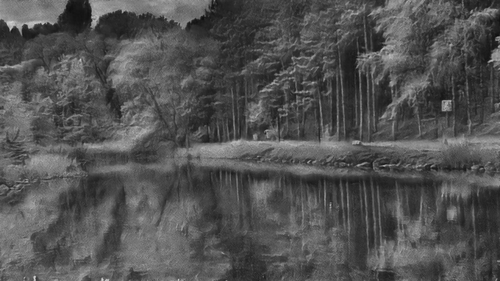} \\	

	\rotatebox[origin=l]{90}{~Night} &
    \includegraphics[width=\widthframes]{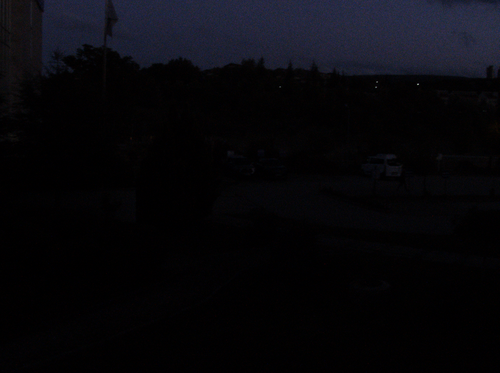} &
	\includegraphics[width=\widthframes]{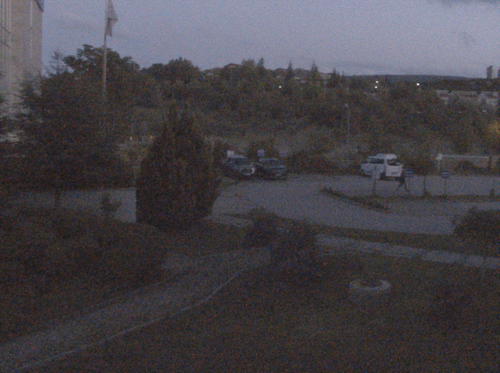} &
	\includegraphics[width=\widthframes]{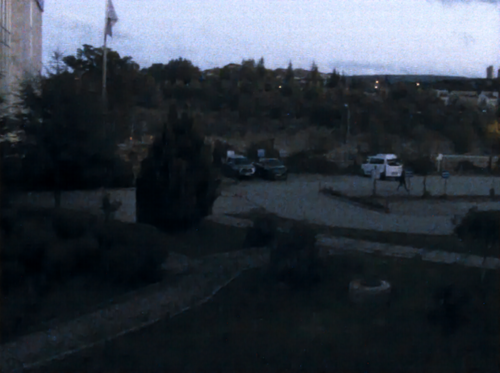} &
	\includegraphics[width=\widthframes]{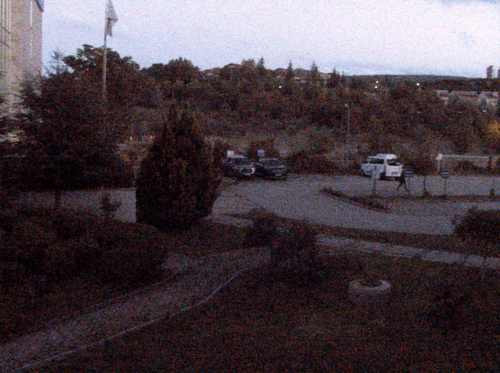} &
	\includegraphics[width=\widthframes]{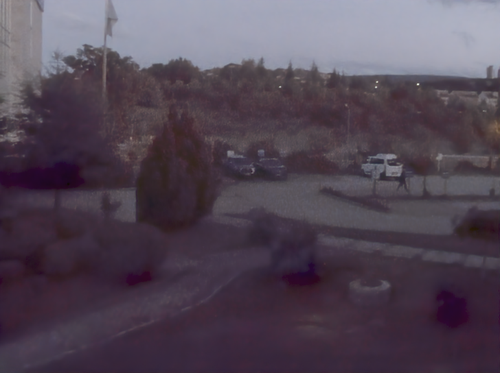} &	
    \includegraphics[width=\widthevents]{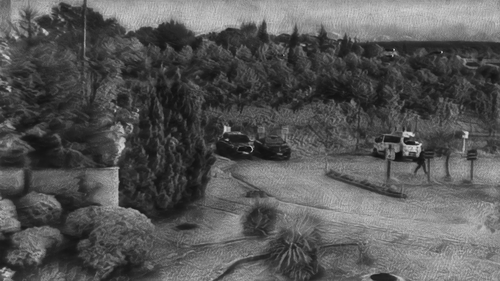} &
	\includegraphics[width=\widthevents]{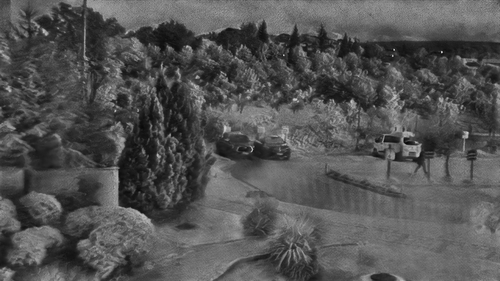} \\

	\rotatebox[origin=l]{90}{Control.} &
    \includegraphics[width=\widthframes]{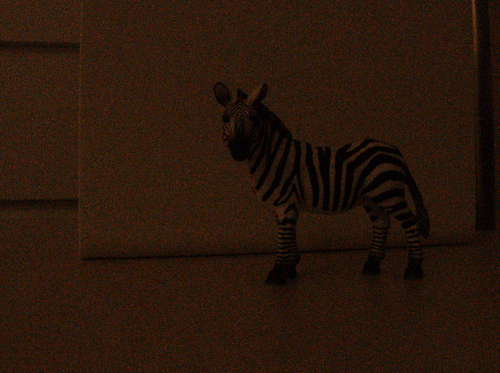} &
	\includegraphics[width=\widthframes]{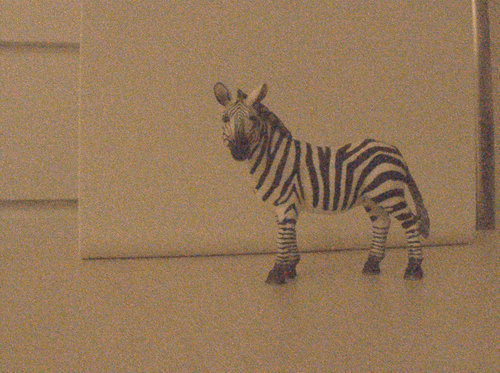} &
	\includegraphics[width=\widthframes]{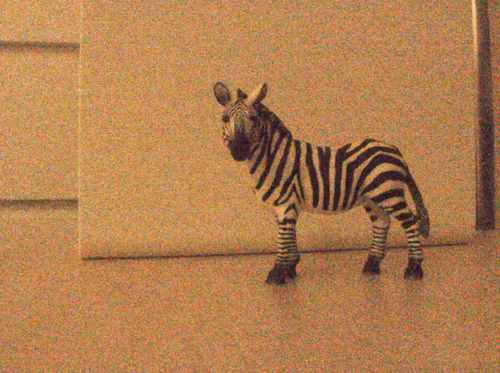} &
	\includegraphics[width=\widthframes]{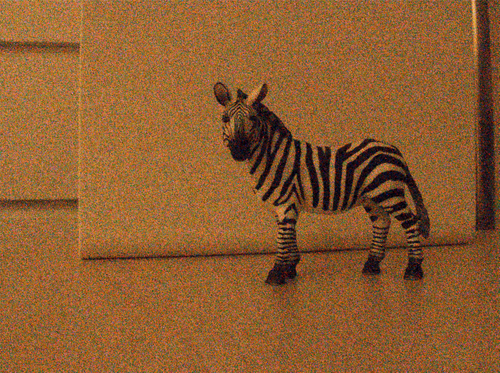} &
	\includegraphics[width=\widthframes]{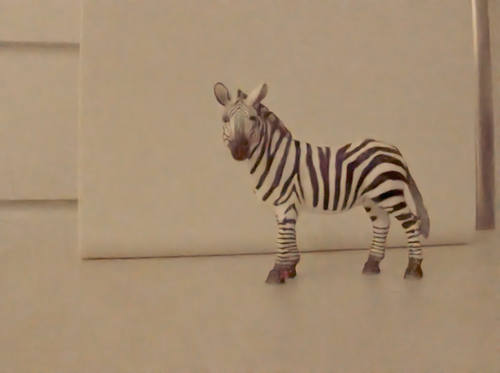} &	
     \includegraphics[width=\widthevents]{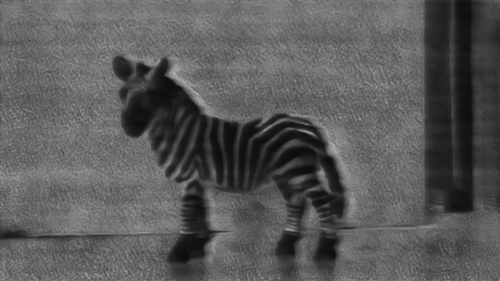} &
	\includegraphics[width=\widthevents]{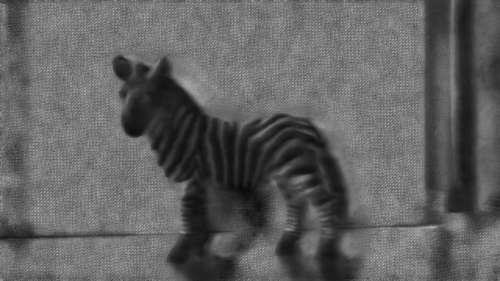} \\

\tiny
    & Input & \makebox[0pt][c]{StableLLVE} & SNR-Net & \makebox[0pt][c]{RetinexF.} & \makebox[0pt][c]{EvLowLight} & ET-Net & HyperE2VID
\end{tabular}
\caption{Qualitative comparison of the low-light enhancement and event-based image reconstruction methods on sample scenes from different subsets of our \datname dataset.}
\vspace{-2mm}
\label{fig:qual_eval}
\end{figure*}

Fig.~\ref{fig:qual_eval} presents qualitative results from sample scenes for each subset of our dataset. The visual results reveal that while ET-Net and HyperE2VID effectively capture texture and structural information, their reconstructions lack color fidelity. Additionally, the high noise levels in low-light conditions are evident in the generated images. EvLowLight demonstrates superior noise reduction compared to RGB-based image and video methods, yet it struggles to maintain accurate color representation. This suggests that the inclusion of event information in EvLowLight significantly aids in noise reduction, resulting in smoother images. Conversely, image and video-based methods, though richer in color, perform less effectively in denoising. In conclusion, based on both qualitative and quantitative results, no single method consistently outperforms the others across all evaluation metrics.

\subsection{Analysis on Downstream Task}
\begin{figure*}[!t]
\scriptsize
\newcommand{\widthframes}{0.24\textwidth}
\newcommand{\widthevents}{0.32\textwidth}
\centering
\setlength{\tabcolsep}{0.3ex} %
\begin{tabular}{cccc}
    \includegraphics[width=\widthframes,trim={0 0 0 200px},clip]{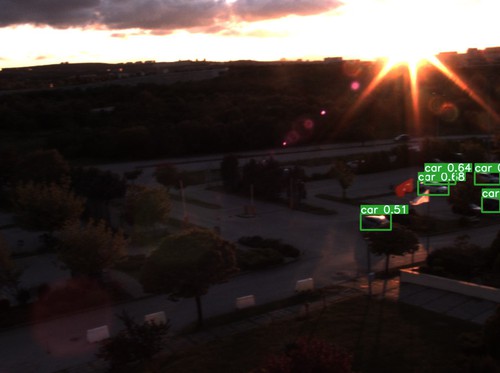} &
    \includegraphics[width=\widthframes,trim={0 0 0 200px},clip]{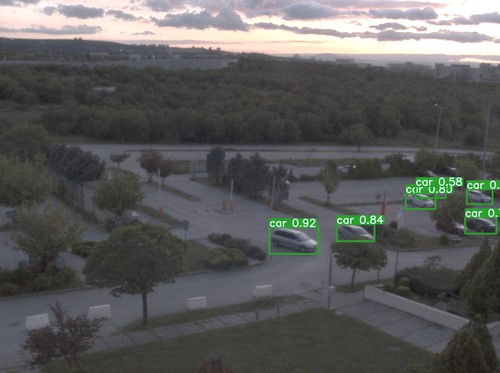} &
    \includegraphics[width=\widthframes,trim={0 0 0 200px},clip]{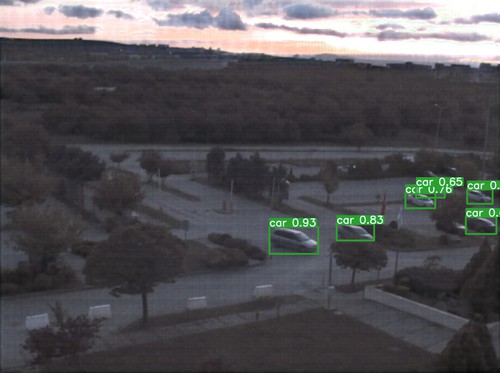} &
    \includegraphics[width=\widthframes,trim={0 0 0 200px},clip]{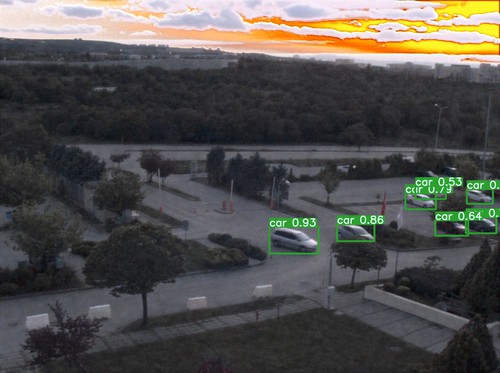} \\

    Input & \makebox[0pt][c]{StableLLVE} & SNR-Net & \makebox[0pt][c]{RetinexFormer} \\
\end{tabular}

\begin{tabular}{ccc}
    \includegraphics[width=\widthframes,trim={0 0 0 200px},clip]{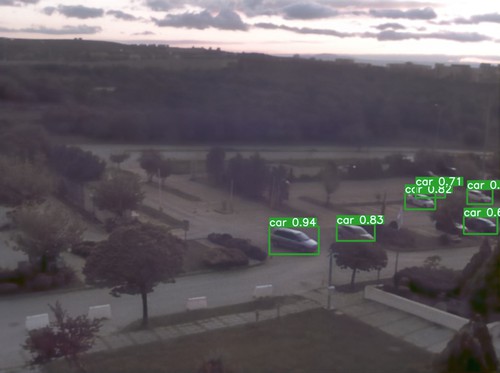} &
    \includegraphics[width=\widthevents,trim={0 0 0 134px},clip]{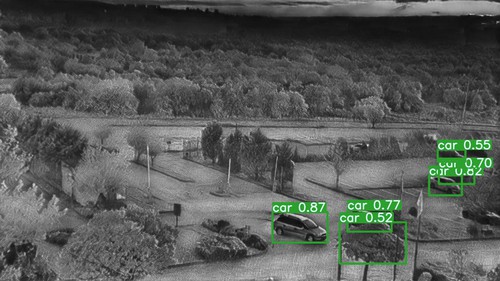} &
    \includegraphics[width=\widthevents,trim={0 0 0 134px},clip]{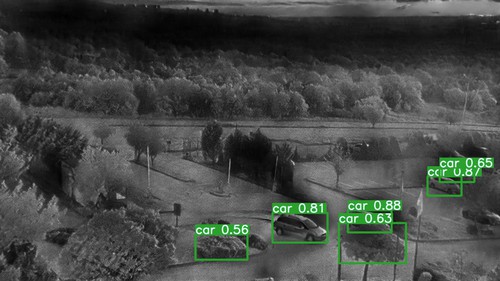} \\

    \makebox[0pt][c]{EvLowLight} & ET-Net & HyperE2VID
\end{tabular}
\caption{Object detection performance on input RGB image, low-light enhancement and event-based image reconstruction methods.}
\vspace{-2mm}
\label{fig:ds_eval}
\end{figure*}

We conducted further evaluations to assess the effectiveness of each method on a downstream task: object detection. Specifically, we used the YOLOv7~\cite{wang2023yolov7} object detector to detect vehicles in the enhanced images produced by low-light enhancement and event-based image reconstruction methods. Fig.~\ref{fig:ds_eval} shows example results from the object detector. Despite the high scores of event-based image reconstruction methods on no-reference metrics, qualitative results reveal that these methods often produce numerous false-positive detections. In contrast, methods such as RetinexFormer accurately detected all vehicles in the scene. These findings demonstrate the importance of color information in downstream tasks like object detection. While event-based methods perform well in certain metrics, they may fall short in practical applications where accurate color representation is crucial.

\section{Conclusion}
In this paper, we introduced the HUE dataset, a comprehensive collection of high-resolution event and frame sequences captured in a wide range of low-light conditions. Our dataset is distinguished by its high resolution, diverse scenarios, and specific focus on challenging low-light environments. It includes sequences recorded both indoors and outdoors, at various times of day, and with differing camera motions and scene dynamics, enabling extensive evaluation of low-light enhancement and event-based image reconstruction methods. 

We assessed several state-of-the-art methods across four categories: RGB image-based, RGB video-based, event-based, and hybrid event-RGB methods. Our quantitative analysis using no-reference metrics revealed that event-based image reconstruction methods generally performed well in metrics like NIQE, MANIQA, and MUSIQ, while RGB-based methods such as RetinexFormer performed best in BRISQUE. However, qualitative evaluations and a downstream object detection task emphasized the importance of accurate color information, with RGB-based methods showing superior performance in detecting objects without false positives. 

These findings highlight the necessity of considering both quantitative metrics and practical application outcomes when evaluating low-light enhancement techniques. While event-based methods are powerful in certain metrics, they may require further refinement for tasks involving color fidelity and object detection. Conversely, hybrid methods like EvLowLight show the potential benefits of integrating event data to reduce noise and improve image smoothness. The \datname dataset, coupled with our detailed evaluations, provides a valuable resource for advancing research in low-light vision and hybrid camera systems. We anticipate that our dataset and findings will inspire future research in this field, ultimately leading to more robust and versatile low-light imaging solutions.
\vspace{-0.05cm}
\section*{Acknowledgments}
This work was supported by TUBITAK-1001 Program Award No. 121E454.

\clearpage  %

\bibliographystyle{splncs04}
\bibliography{ms}
\end{document}


\title{\datname Dataset: High-Resolution Event and Frame Sequences for Low-Light Vision \\ Supplementary Material} 

\titlerunning{HUE Dataset}

\author{Burak Ercan$^*$\inst{1}\orcidlink{0000-0002-9231-7982} \and
Onur Eker$^*$\inst{1,2}\orcidlink{0000-0003-4040-6438} \and
Aykut Erdem\inst{3,4}\orcidlink{0000-0002-6280-8422} \and
Erkut Erdem\inst{1}\orcidlink{0000-0002-6744-8614}}

\authorrunning{B.~Ercan et al.}

\institute{Hacettepe University, Computer Engineering Department \and
HAVELSAN Inc. \and
Ko\c{c} University, Computer Engineering Department \and
Ko\c{c} University, KUIS AI Center
}

\maketitle
\def\thefootnote{*}\footnotetext{These authors contributed equally to this work}

In this supplementary document, we provide a breakdown of all the sequences in our dataset. We give a separate table for each of the dataset categories, where each row in a table stands for a single sequence. For each sequence, we give the name of the sequence (column 1), duration of the sequence in seconds (column 2 - Dur.), number of frames in the sequence (column 3 - Fr.), number of events in the sequence in millions (column 4 - Ev.(M)), sensor illuminance level in lux (column 5), exposure time of the frame camera (column 6 - Exp.), digital gain setting of the frame camera (column 7), characteristic scene motion in the sequence (column 8 - SM), dominant light source in the scene (column 9 - LS), and a short description of the sequence (column 10). For column 8, D stands for dynamic scenes and S stands for static scenes. For column 9, N stands for natural lights and A stands for artificial lights. In \Cref{tab:hue_seqs_indoor,tab:hue_seqs_city,tab:hue_seqs_twilight,tab:hue_seqs_night,tab:hue_seqs_driving,tab:hue_seqs_controlled} we present these for Indoor, City, Twilight, Night, Driving, and Controlled categories of our dataset, respectively.

\begin{landscape} 
\begin{table}[htbp]
\caption{Breakdown of sequences in \datname-Indoor.}
\centering\scriptsize
\renewcommand{\arraystretch}{1.2}
\begin{tabular}{L{3cm} C{1cm} C{1cm} C{1.2cm} C{1cm} C{0.8cm} C{0.9cm} C{0.6cm} C{0.6cm} C{0.1cm} L{7cm}}
\toprule
\textbf{Name}  & \textbf{Dur.} & \textbf{Fr.} & \textbf{Ev.(M)} & \textbf{Lux} & \textbf{Exp.} & \textbf{Gain} & \textbf{SM} & \textbf{LS} && \textbf{Description} \\
\midrule
\texttt{bookshelves} & 14,4 & 361 & 542,94 & 0 & 35 & 15 & S & N && Indoor bookshelves \\
\texttt{building\_entrance} & 18,6 & 465 & 404,83 & 0 & 35 & 30 & S & N && Entrance of building, board \\
\texttt{corridor\_person} & 26,1 & 654 & 593,02 & 3 & 35 & 10 & D & N && Corridor, walk, window view and a dynamic person  \\
\texttt{corridor\_selfie} & 10,8 & 271 & 244,36 & 2 & 35 & 10 & D & N && Indoor, corridor, mostly rotation, and selfie \\
\texttt{dolls} & 12,4 & 309 & 172,92 & 0 & 35 & 15 & S & N && Indoor dolls \\
\texttt{dome} & 7,4 & 184 & 115,73 & 0 & 35 & 30 & S & A && Dome of the hall and writings on it \\
\texttt{figures\_classics} & 18,7 & 467 & 628,84 & 0 & 35 & 15 & S & N && Shelves, figures and CD collection \\
\texttt{frames} & 17,8 & 446 & 387,91 & 0 & 35 & 15 & S & N && Indoor photo frames and wall decoration \\
\texttt{lab\_1} & 11,9 & 299 & 536,04 & 1 & 30 & 15 & S & N && Laboratory, static scene \\
\texttt{lab\_2} & 4,9 & 123 & 194,05 & 0 & 35 & 15 & S & N && Laboratory, static scene \\
\texttt{lab\_3} & 34,7 & 869 & 1340,72 & 1 & 35 & 30 & S & N && Laboratory, static scene, longer \\
\texttt{letters} & 9,4 & 234 & 355,07 & 0 & 35 & 30 & S & A && Display of hanging letters in the public library \\
\texttt{miniature} & 8,8 & 220 & 156,21 & 0 & 35 & 30 & D & A && Display of old miniature paintings \\
\texttt{old\_books} & 8,4 & 211 & 134,29 & 0 & 35 & 30 & D & A && Display of old books \\
\texttt{old\_classroom\_1} & 17,7 & 442 & 254,37 & 0 & 35 & 30 & S & A && Old classroom \\
\texttt{old\_classroom\_2} & 17,1 & 427 & 259,29 & 0 & 35 & 40 & S & A && Old classroom \\
\texttt{posters\_window} & 32,1 & 803 & 1050,20 & 0 & 35 & 30 & D & N && Posters and window view walking people \\
\texttt{recycle\_art} & 7,2 & 179 & 133,79 & 0 & 35 & 30 & S & N && Recycle art of birds, writing \\
\texttt{selfie} & 5,0 & 125 & 54,56 & 0 & 35 & 20 & D & N && Indoor selfie \\
\texttt{stairs\_dark} & 30,1 & 754 & 244,51 & 0 & 35 & 40 & S & N && Going down the stairs, dark, shaky \\
\texttt{stairways} & 11,5 & 287 & 215,64 & 0 & 35 & 30 & D & A && Stairways of large public library, people walking \\
\texttt{very\_dark\_hand} & 3,8 & 96 & 22,82 & 0 & 35 & 5 & D & A && Very dark room, hand motion \\
\texttt{very\_dark\_room} & 7,6 & 189 & 34,73 & 0 & 35 & 5 & S & A && Very dark room, camera motion \\
\bottomrule
\end{tabular}
\label{tab:hue_seqs_indoor}
\end{table}
\end{landscape}

\begin{landscape} 
\begin{table}[htbp]
\caption{Breakdown of sequences in \datname-City.}
\centering\scriptsize
\renewcommand{\arraystretch}{1.2}
\begin{tabular}{L{3cm} C{1cm} C{1cm} C{1.2cm} C{1cm} C{0.8cm} C{0.9cm} C{0.6cm} C{0.6cm} C{0.1cm} L{7cm}}
\toprule
\textbf{Name}  & \textbf{Dur.} & \textbf{Fr.} & \textbf{Ev.(M)} & \textbf{Lux} & \textbf{Exp.} & \textbf{Gain} & \textbf{SM} & \textbf{LS} && \textbf{Description} \\
\midrule
\texttt{city\_night\_1} & 13,3 & 332 & 60,41 & 0 & 35 & 48 & S & A && City view from window, very dark night \\
\texttt{city\_night\_2} & 12,6 & 316 & 36,34 & 0 & 35 & 48 & S & A && City view from window, very dark night \\
\texttt{hdr\_plants\_1} & 21,2 & 530 & 606,94 & 12 & 35 & 10 & S & N && Indoor plants and window city view, HDR \\
\texttt{hdr\_plants\_2} & 20,6 & 516 & 601,89 & 14 & 35 & 10 & S & N && Indoor plants and window city view, HDR \\
\texttt{city\_twilight\_1} & 14,4 & 360 & 240,98 & 4 & 35 & 10 & S & N && Window view of city in twilight \\
\texttt{city\_twilight\_2} & 21,9 & 549 & 372,87 & 1 & 35 & 10 & D & N && Window view in twilight, close by car  \\
\texttt{city\_twilight\_3} & 18,4 & 461 & 178,79 & 5 & 35 & 10 & S & N && Window view of city in twilight \\
\texttt{city\_twilight\_4} & 25,5 & 639 & 572,84 & 1 & 35 & 10 & S & N && Window view of trees and apartments  \\
\texttt{city\_twilight\_5} & 22,3 & 557 & 447,07 & 1 & 35 & 10 & S & N && Window view of closer cars and apartments  \\
\texttt{city\_twilight\_6} & 23,1 & 577 & 362,18 & 0 & 35 & 10 & D & N && Window view in twilight, two people walk  \\
\texttt{plants\_and\_city} & 22,9 & 572 & 388,80 & 0 & 35 & 30 & S & N && Low light İndoor plants and window view \\
\bottomrule
\end{tabular}
\label{tab:hue_seqs_city}
\end{table}
\end{landscape}

\begin{landscape} 
\begin{table}[htbp]
\caption{Breakdown of sequences in \datname-Twilight.}
\centering\scriptsize
\renewcommand{\arraystretch}{1.2}
\begin{tabular}{L{3.6cm} C{1cm} C{1cm} C{1.2cm} C{1cm} C{0.8cm} C{0.9cm} C{0.6cm} C{0.6cm} C{0.1cm} L{7cm}}
\toprule
\textbf{Name}  & \textbf{Dur.} & \textbf{Fr.} & \textbf{Ev.(M)} & \textbf{Lux} & \textbf{Exp.} & \textbf{Gain} & \textbf{SM} & \textbf{LS} && \textbf{Description} \\
\midrule
\texttt{dark\_equipment\_1} & 7,4 & 186 & 57,78 & 0 & 35 & 35 & S & N && Dark forest and construction equipment  \\
\texttt{dark\_equipment\_2} & 20,3 & 507 & 163,92 & 0 & 35 & 45 & S & N && Forest, path, and construction equipment  \\
\texttt{dark\_forest\_1} & 30,4 & 760 & 551,10 & 0 & 35 & 36 & S & N && Dark forest \\
\texttt{dark\_forest\_2} & 11,3 & 282 & 177,69 & 0 & 35 & 36 & S & N && Dark forest and illuminated path ahead \\
\texttt{duck\_fence} & 15,1 & 379 & 266,56 & 0 & 35 & 32 & D & N && Ducks behind fence,  sign with writings \\
\texttt{duck\_fence\_lake} & 22,7 & 569 & 811,88 & 0 & 35 & 32 & D & N && Lake and ducks behind fences \\
\texttt{duck\_lake} & 38,9 & 974 & 1322,53 & 1 & 35 & 20 & D & N && Ducks standing and walking and bathing, waves \\
\texttt{duck\_lake\_2} & 28,8 & 721 & 849,72 & 0 & 35 & 32 & D & N && Lake with small waves, two ducks stand and move \\
\texttt{duck\_lake\_3} & 35,6 & 891 & 552,69 & 0 & 35 & 32 & D & N && Ducks swimming in the lake \\
\texttt{duck\_lake\_pause} & 34,2 & 855 & 259,18 & 0 & 35 & 30 & D & N && Ducks swimming, camera stationary at the end \\
\texttt{hdr\_terrace\_sun\_1} & 32,3 & 807 & 1032,75 & 24 & 30 & 10 & D & N && Terrace, walking, reflections, sun, HDR, car \\
\texttt{lake\_1} & 13,6 & 340 & 310,04 & 1 & 35 & 30 & S & N && Cafe, lake with reflections, people across the lake \\
\texttt{lake\_2} & 14,8 & 370 & 275,57 & 1 & 35 & 10 & S & N && Cafe, lake with reflections, underexposed frames \\
\texttt{lake\_3} & 23,8 & 595 & 426,14 & 1 & 35 & 15 & S & N && Cafe, lake with reflections, mostly static scene \\
\texttt{lake\_turtle} & 24,1 & 603 & 222,70 & 0 & 35 & 20 & D & N && A barely visible turtle swimming in lake \\
\texttt{sunset\_parking\_lot} & 24,6 & 615 & 408,38 & 3 & 35 & 10 & D & N && Parking lot, sun is down, a few cars pass \\
\texttt{sunset\_parking\_lot\_pause\_1} & 40,2 & 1005 & 614,94 & 3 & 35 & 10 & D & N && Parking lot, camera stationary at the end \\
\texttt{sunset\_parking\_lot\_pause\_2} & 23,4 & 585 & 249,12 & 1 & 35 & 10 & D & N && Parking lot, camera stationary at the end \\
\texttt{terrace\_flies} & 37,1 & 929 & 402,61 & 2 & 30 & 20 & D & N && Getting darker, on the terrace, flies \\
\texttt{terrace\_puddle} & 19,1 & 477 & 325,49 & 0 & 30 & 20 & S & N && Walking besides puddle on the terrace, reflections \\
\texttt{terrace\_sunset} & 21,9 & 548 & 715,16 & 5 & 35 & 10 & D & N && Terrace, laptop, reflection selfie, sunset \\
\texttt{waterflow\_1} & 25,9 & 647 & 786,63 & 1 & 35 & 25 & D & N && Water flowing, long trees \\
\texttt{waterflow\_2} & 26,5 & 662 & 488,97 & 0 & 35 & 32 & D & N && Water flowing, camera slowly moves \\
\bottomrule
\end{tabular}
\label{tab:hue_seqs_twilight}
\end{table}
\end{landscape}

\begin{landscape} 
\begin{table}[htbp]
\caption{Breakdown of sequences in \datname-Night.}
\centering\scriptsize
\renewcommand{\arraystretch}{1.2}
\begin{tabular}{L{3.6cm} C{1cm} C{1cm} C{1.2cm} C{1cm} C{0.8cm} C{0.9cm} C{0.6cm} C{0.6cm} C{0.1cm} L{7cm}}
\toprule
\textbf{Name}  & \textbf{Dur.} & \textbf{Fr.} & \textbf{Ev.(M)} & \textbf{Lux} & \textbf{Exp.} & \textbf{Gain} & \textbf{SM} & \textbf{LS} && \textbf{Description} \\
\midrule
\texttt{night\_face} & 5,8 & 145 & 24,02 & 0 & 35 & 48 & D & A && Person face (further) \\
\texttt{night\_face\_close} & 7,8 & 196 & 63,58 & 0 & 35 & 48 & D & A && Person face (close) \\
\texttt{night\_park\_1} & 5,8 & 146 & 12,62 & 0 & 35 & 48 & S & A && Dark, trees, buildings far away \\
\texttt{night\_park\_2} & 5,1 & 128 & 7,36 & 0 & 35 & 48 & S & A && Park \\
\texttt{night\_park\_3} & 9,6 & 241 & 66,62 & 0 & 35 & 48 & S & A && Park, closer to lamp, brighter \\
\texttt{night\_park\_4} & 6,9 & 173 & 9,30 & 0 & 35 & 48 & S & A && Park \\
\texttt{night\_park\_person\_1} & 10,2 & 255 & 25,30 & 0 & 35 & 48 & D & A && Park, person walking \\
\texttt{night\_park\_person\_2} & 12,5 & 313 & 37,82 & 0 & 35 & 48 & D & A && Park, person walking \\
\texttt{night\_park\_swing} & 8,6 & 216 & 3,80 & 0 & 35 & 48 & D & A && Park, swinging \\
\texttt{night\_park\_walk\_1} & 11,1 & 278 & 36,22 & 0 & 35 & 48 & S & A && Park, walking, light and dark regions \\
\texttt{night\_park\_walk\_2} & 12,5 & 313 & 64,01 & 0 & 35 & 48 & S & A && Park, walking and panning \\
\texttt{night\_parking\_lot} & 23,9 & 598 & 148,11 & 0 & 35 & 30 & D & A && Camera moves slowly, a person walks \\
\texttt{night\_street\_1} & 11,4 & 285 & 46,43 & 0 & 35 & 48 & S & A && Streets, buildings \\
\texttt{night\_street\_car\_1} & 5,8 & 146 & 24,59 & 0 & 35 & 48 & D & A && Car passing, person \\
\texttt{night\_street\_car\_2} & 2,9 & 73 & 9,16 & 0 & 35 & 48 & D & A && Car passing \\
\texttt{night\_street\_car\_bike} & 13,1 & 328 & 63,07 & 0 & 35 & 48 & D & A && Three car and a motorbike passing \\
\bottomrule
\end{tabular}
\label{tab:hue_seqs_night}
\end{table}
\end{landscape}

\begin{landscape} 
\begin{table}[htbp]
\caption{Breakdown of sequences in \datname-Driving.}
\centering\scriptsize
\renewcommand{\arraystretch}{1.2}
\begin{tabular}{L{1.6cm} C{1cm} C{1cm} C{1.2cm} C{1cm} C{0.8cm} C{0.9cm} C{0.6cm} C{0.6cm} C{0.1cm} L{9cm}}
\toprule
\textbf{Name}  & \textbf{Dur.} & \textbf{Fr.} & \textbf{Ev.(M)} & \textbf{Lux} & \textbf{Exp.} & \textbf{Gain} & \textbf{SM} & \textbf{LS} && \textbf{Description} \\
\midrule
\texttt{drive\_1} & 20,1 & 502 & 249,56 & 0 & 35 & 30 & D & N && Driving in the street, a few cars moving ahead  \\
\texttt{drive\_2} & 23,1 & 577 & 157,05 & 0 & 35 & 30 & D & N && Camera moves slowly inside a car at the intersection \\
\texttt{drive\_3} & 27,0 & 675 & 181,50 & 0 & 35 & 30 & D & N && Driving on the road, other cars moving in the same or opposite direction \\
\texttt{drive\_4} & 9,6 & 240 & 85,13 & 0 & 35 & 30 & D & N && Driving on the road, a moving pedestrian and car ahead   \\
\texttt{drive\_5} & 30,4 & 760 & 195,78 & 0 & 35 & 30 & D & N && Driving faster on a three lane road, under a bridge, other cars moving \\
\texttt{drive\_6} & 20,6 & 515 & 691,15 & 0 & 35 & 30 & D & N && Driving and stopping at the intersection, other cars moving \\
\texttt{drive\_7} & 38,3 & 958 & 354,44 & 0 & 35 & 30 & D & N && Driving through the interchange and roundabout, other cars moving \\
\texttt{drive\_8} & 48,2 & 1205 & 262,87 & 0 & 35 & 30 & D & N && Driving on the road, lots of vehicles moving in the opposite direction \\
\texttt{drive\_9} & 19,9 & 498 & 141,99 & 0 & 35 & 30 & D & N && Driving on a very dark street, passing by a pedestrian and a bus \\
\texttt{drive\_10} & 31,2 & 780 & 209,17 & 0 & 35 & 30 & D & N && Driving and turning in dark streets \\
\texttt{drive\_11} & 6,0 & 151 & 67,96 & 0 & 35 & 40 & D & A && Driving in the street, cars moving in the opposite direction \\
\texttt{drive\_12} & 9,6 & 239 & 42,68 & 0 & 35 & 40 & D & A && Driving on the road and under a signboard illuminated by headlights \\
\texttt{drive\_13} & 16,0 & 401 & 61,47 & 0 & 35 & 40 & D & A && Driving on a four lane road, under signboards illuminated intermittently   \\
\texttt{drive\_14} & 22,5 & 563 & 113,97 & 0 & 35 & 40 & D & A && Driving at dark road, roundabout, just a few other cars \\
\texttt{drive\_15} & 19,4 & 486 & 68,07 & 0 & 35 & 40 & D & A && Driving at mostly straight dark road, few other cars \\
\texttt{drive\_16} & 21,1 & 527 & 97,87 & 0 & 35 & 45 & S & A && Driving at mostly straight very dark road, no other cars \\
\bottomrule
\end{tabular}
\label{tab:hue_seqs_driving}
\end{table}
\end{landscape}

\begin{landscape} 
\begin{table}[htbp]
\caption{Breakdown of sequences in \datname-Controlled.}
\centering\scriptsize
\renewcommand{\arraystretch}{1.2}
\begin{tabular}{L{2.6cm} C{1cm} C{1cm} C{1.2cm} C{1cm} C{0.8cm} C{0.9cm} C{0.6cm} C{0.6cm} C{0.1cm} L{7cm}}
\toprule
\textbf{Name}  & \textbf{Dur.} & \textbf{Fr.} & \textbf{Ev.(M)} & \textbf{Lux} & \textbf{Exp.} & \textbf{Gain} & \textbf{SM} & \textbf{LS} && \textbf{Description} \\
\midrule
\texttt{zebra\_L1\_G10} & 7,2 & 180 & 29,59 & 2 & 35 & 10 & S & A && Static toy zebra at light level L1 (brightest) \\
\texttt{zebra\_L1\_G6} & 10,7 & 267 & 37,03 & 2 & 35 & 6 & S & A && Static toy zebra at light level L1 \\
\texttt{zebra\_L2\_G16} & 10,3 & 257 & 34,26 & 1 & 35 & 16 & S & A && Static toy zebra at light level L2 \\
\texttt{zebra\_L2\_G12} & 7,1 & 177 & 28,51 & 1 & 35 & 12 & S & A && Static toy zebra at light level L2 \\
\texttt{zebra\_L3\_G23} & 7,7 & 192 & 22,15 & 0 & 35 & 23 & S & A && Static toy zebra at light level L3 \\
\texttt{zebra\_L3\_G19} & 6,0 & 149 & 21,44 & 0 & 35 & 19 & S & A && Static toy zebra at light level L3 \\
\texttt{zebra\_L4\_G27} & 7,4 & 186 & 17,54 & 0 & 35 & 27 & S & A && Static toy zebra at light level L4 \\
\texttt{zebra\_L4\_G22} & 7,2 & 180 & 16,34 & 0 & 35 & 22 & S & A && Static toy zebra at light level L4 \\
\texttt{zebra\_L5\_G34} & 7,8 & 194 & 13,77 & 0 & 35 & 34 & S & A && Static toy zebra at light level L5 \\
\texttt{zebra\_L5\_G29} & 6,8 & 171 & 17,19 & 0 & 35 & 29 & S & A && Static toy zebra at light level L5 \\
\texttt{zebra\_L6\_G39} & 8,9 & 222 & 12,60 & 0 & 35 & 39 & S & A && Static toy zebra at light level L6 \\
\texttt{zebra\_L6\_G35} & 5,8 & 146 & 11,60 & 0 & 35 & 35 & S & A && Static toy zebra at light level L6 \\
\texttt{zebra\_L7\_G46} & 6,9 & 172 & 7,33 & 0 & 35 & 46 & S & A && Static toy zebra at light level L7 \\
\texttt{zebra\_L7\_G42} & 7,4 & 184 & 7,11 & 0 & 35 & 42 & S & A && Static toy zebra at light level L7 \\
\texttt{zebra\_L8\_G48} & 6,6 & 164 & 2,26 & 0 & 35 & 48 & S & A && Static toy zebra at light level L8 \\
\texttt{zebra\_L9\_G48} & 6,6 & 164 & 1,61 & 0 & 35 & 48 & S & A && Static toy zebra at light level L9 \\
\texttt{zebra\_L10\_G48} & 8,6 & 214 & 0,99 & 0 & 35 & 48 & S & A && Static toy zebra at light level L10 (darkest) \\
\bottomrule
\end{tabular}
\label{tab:hue_seqs_controlled}
\end{table}
\end{landscape}